\documentclass{article} 
\usepackage[preprint]{colm2026_conference}

\usepackage{microtype}
\usepackage{hyperref}
\usepackage{url}
\usepackage{booktabs}
\usepackage{graphicx}
\usepackage{amsmath}
\usepackage{multirow}
\usepackage{longtable}
\usepackage{booktabs}
\usepackage{array}
\usepackage{makecell}
\usepackage[table]{xcolor}

\usepackage{comment} 
\usepackage{lineno}
\usepackage{subcaption}
\usepackage{color-edits}
\addauthor[Mike]{mf}{purple}
\addauthor[Steven]{sf}{blue}

\definecolor{darkblue}{rgb}{0, 0, 0.5}
\hypersetup{colorlinks=true, citecolor=darkblue, linkcolor=darkblue, urlcolor=darkblue}

\newenvironment{tight_itemize}{
\begin{itemize}
  \setlength{\itemsep}{0pt}
  \setlength{\parskip}{0pt}
}{\end{itemize}}

\newcommand\blfootnote[1]{%
  \begingroup
  \renewcommand\thefootnote{}\footnote{#1}%
  \addtocounter{footnote}{-1}%
  \endgroup
}

\title{
Baby Scale: Investigating Models Trained on Individual Children's Language Input
}

\author{Steven Y. Feng, Alvin W.M. Tan, Michael C. Frank \\
Stanford University\\
\texttt{\{syfeng,tanawm,mcfrank\}@stanford.edu} \\
}

\begin{document}

\ifcolmsubmission
\linenumbers
\fi

\maketitle

\begin{abstract}

Modern language models (LMs) must be trained on many orders of magnitude more words of training data than human children receive before they begin to produce useful behavior. 
Assessing the nature and origins of this ``data gap'' requires benchmarking LMs on human-scale datasets to understand how linguistic knowledge emerges from children's natural training data. 
Using transcripts from the BabyView dataset (videos from children ages 6--36 months), we investigate (1) scaling performance at child-scale data regimes, (2) variability in model performance across datasets from different children's experiences and linguistic predictors of dataset quality, and (3) relationships between model and child language learning outcomes. 
LMs trained on child data show acceptable scaling for grammar tasks, but lower scaling on semantic and world knowledge tasks than models trained on synthetic data; we also observe substantial variability on data from different children. Beyond dataset size, performance is most associated with a combination of distributional and interactional linguistic features, broadly consistent with what makes high-quality input for child language development. 
Finally, model likelihoods for individual words correlate with children’s learning of those words, suggesting that properties of child-directed input may influence both model learning and human language development. 
Overall, understanding what properties make language data efficient for learning can enable more powerful small-scale language models while also shedding light on human language acquisition.\blfootnote{Code and data: \url{https://github.com/styfeng/babyscale-LM}} 


\end{abstract}

\section{Introduction}\label{sec:intro}

Large language models (LLMs) achieve impressive capabilities by training on enormous datasets consisting of billions or trillions of tokens. In contrast, human children acquire language with orders of magnitude less linguistic input. Children receive between $10^7$ and $10^8$ words during early development -- far smaller than the datasets used to train modern LLMs, which are often $10^{12}$--$10^{13}$ words \citep{frank2023, warstadt2023findings}. This ``data gap'' likely results from many causes, including language model architecture, the nature of their training, and the nature of the data they are trained on. Understanding what makes children more efficient learners -- and potentially importing these insights to language models -- requires the evaluation of language models on datasets from human language acquisition. 

Small language models (SLMs) trained on human-scale datasets have emerged as an important tool for investigating data efficiency. For example, the BabyLM challenges constrain training data to developmentally plausible budgets and ask participants to develop architectures and training methodologies that learn efficiently under such conditions \citep{warstadt2023findings, hu2024findings}. These competitions have led to substantial advances \citep[e.g.,][]{charpentier2024gpt, martinez-etal-2023-climb}. Yet, by holding the training data constant -- a necessary feature of these challenges -- they prevent researchers from answering an important but distinct set of questions about what makes some language datasets more efficient for learning than others, and to what extent efficient learning might be driven by certain types of data.

Here, we use data from children's learning environments to study what makes data ``good" for learning at human scales. We are inspired by developmental psychology studies of the importance of dataset quantity vs. quality in children's language learning \citep{anderson2021linking,masek2021beyond, rowe2012longitudinal, rowe2020analyzing}, which suggest that children learn more from input that contains complex sentence structures and diverse vocabulary.

Data are a key bottleneck in studying child language. Although several tens of millions of words of data from children's naturalistic conversations are available in the CHILDES archive \citep{macwhinney2014childes}, it is challenging to compare across them as they are collected from very different studies with different methods. Further, these corpora typically do not have outcome measures on language acquisition in the individual children who contribute, meaning that the datasets cannot be linked to individual children's learning progress. We address these issues by using the BabyView dataset \citep{long2024babyview}, a collection of egocentric recordings (with transcripts) capturing naturalistic language experiences by children across multiple families. Because each family represents a distinct linguistic environment and context for language learning, BabyView enables us to train SLMs on individual families' data to simulate "synthetic learners" exposed to different language learning environments \citep[following][]{wang2023finding,qin2024systematic}. 

Our setting allows us to systematically examine several key research questions: (1) How does language model performance scale with the amount of input received by a single child, or a mixture of children? (2) How much variability arises across different families' language environments, and which linguistic features of the input data predict language model performance?
(3) Do properties of child-directed input relate to children's language outcomes?
Our results reveal that scaling performance varies across tasks: while scaling remains visible in grammar even at child-scale data regimes, semantic and world knowledge tasks scale much less well. Further, variability across families is substantial and explained by factors beyond dataset size, including distributional and interactional properties. Finally, model likelihoods relate to individual children's outcomes. Together, these results provide new insight into the role of input data in language learning and highlight the potential of child-scale datasets as a lens for understanding human and machine language acquisition.

\section{Background and Related Work}

Scaling laws have become a central framework for understanding LM performance. Prior work demonstrates that performance often follows power-law relationships with model size, dataset size, and compute. However, most empirical studies examine regimes involving billions of tokens \citep{kaplan2020scalinglawsneurallanguage,henighan2020scalinglawsautoregressivegenerative,hoffmann2022training,muennighoff2023scaling}. Further, there is a body of work on optimal data mixing and selection \citep{NEURIPS2023_a8f8cbd7,gao2020pile800gbdatasetdiverse,10.5555/3618408.3619349,feng2024maximizedataspotentialenhancing}, but also at much larger scales. While there are studies examining how scaling laws vary based on the data being provided \citep{NEURIPS2022_7b75da9b,pandey2024gzippredictsdatadependentscaling,brill2024neuralscalinglawsrooted}, much less is known about scaling behavior in extremely small data regimes. Understanding whether similar scaling dynamics emerge at child-scale data sizes -- and how they are modulated by aspects of dataset quality -- can provide insight into both model efficiency and the role of input data.

Much recent work on the properties of SLMs comes from the BabyLM challenge \citep{warstadt2023findings,hu2024findings}, which investigates SLM training using fixed budgets of 10M and 100M tokens in a heterogeneous mix of child-directed speech with other text data. Other work has used this kind of paradigm to understand questions related to dataset ``quality'' \citep{zhang2020you, huebner2021babyberta} -- often by using a ``controlled rearing'' design in which models are held constant while varying datasets \citep{frank2023openly}. These designs allow inferences about models' inductive biases \citep{kallini2024mission}, systematic asymmetries \citep{hu2025production}, their ability to generalize to unseen grammatical structures \citep{misra2024language}, and to learn from multilingual input \citep{constantinescu-etal-2025-investigating}, among other topics. 

Our work uses this design to investigate the informational value of child-directed speech. Initial evidence suggested that a pure corpus of child-directed speech might be \emph{less} informative than the BabyLM mixture, 
but this study did not provide evidence for why 
\citep{feng-etal-2024-child}. \citet{qin2024systematic} took important steps by training a range of SLMs on data from three children. However, the datasets being compared were from very different studies and corpora, and no attempt was made to interpret between-child differences.

In fact, a rich literature on children's language development suggests that there are substantial differences in language acquisition between children \citep{frank2021variability, kidd2020individual}, and that these differences are related to differences in children's language input \citep{hart1997meaningful,weisleder2013talking,coffey2025strong}. While initial evidence on this question focused on differences in the \emph{quantity} of speech that children hear across households \citep{hart1997meaningful}, subsequent investigations have also emphasized that other measures of linguistic \emph{quality} -- including lexical and syntactic diversity -- may be even more important \citep{masek2021beyond, anderson2021linking, rowe2012longitudinal, rowe2020analyzing}.

In sum, our work here builds on prior work investigating ``controlled rearing'' in SLMs to understand aspects of dataset quality at child-scale. Inspired by research on child language acquisition, and using dense longitudinal transcripts from a single study, we analyze the features that relate to the effectiveness of child-directed speech as training data.

\section{Methods}

\subsection{Datasets}

Our training data consists of conversation between children and caregivers, set up so that each line corresponds to a single conversation. Each training dataset was split into 85/15 train/val splits. Examples and preprocessing details can be found in Appendix \ref{app:dataset_details}.

{\bf BabyView.} The BabyView dataset consists of egocentric video recordings capturing children’s everyday experiences at home \citep{long2024babyview}. The 2025.2 release of the dataset includes $\sim1500$ hours of video, speech transcripts using WhisperX large-v3 \citep{radford2023robust}, and speaker role diarization using VTC 2.0 \citep{kunze2025challenges}. For this study, we use data only from 20 monolingual English-speaking families ($BV_{Eng}$), identified through a combination of automatic and manual filtering. These individual family transcripts range from $\sim$1000 to $\sim$725k tokens, with an aggregate total of $\sim$2.8M tokens in $BV_{Eng}$.

{\bf TinyDialogues.} The total amount of data in $BV_{Eng}$ is very limited, but this limitation reflects the overall scarcity of child-directed speech transcripts; the CHILDES archive contains only $\sim$29M English words \citep{huebner2021babyberta,macwhinney2014childes}. Thus, to investigate further scaling properties, we also fit models on varying subsets of a larger dataset (up to 200M words) of synthetic child-directed dialogues, the TinyDialogues corpus \citep{feng-etal-2024-child}. Inspired by the TinyStories project \citep{eldan2023tinystories}, these dialogues are generated by GPT-4 to feature diverse situations and vocabulary in short, simple dialogues between children (of varying ages) and caregivers. 

\subsection{Experimental Conditions}\label{subsec:exp_conditions}

We evaluate several different training conditions, totaling 25 experiments:

\begin{tight_itemize}
\item \textbf{Individual-family models}: one model trained per individual $BV_{Eng}$ family (20).
\item \textbf{Pooled mixtures}: models trained on size-controlled mixtures of differing numbers of $BV_{Eng}$ families (2, 5, 10, and all 20), matching the total linguistic data ($\sim$725k tokens) of the largest individual $BV_{Eng}$ family. We proportionally sample conversations from the individual families to most closely reach the desired token count. 
\item \textbf{All-families model}: trained on the entirety (all families) of $BV_{Eng}$.
\end{tight_itemize}

These conditions allow us to compare the effects of child-scale dataset size and diversity on language model learning.


\subsection{Model Architectures}\label{subsec:model}

We train two main model families. First, we train the autoregressive LM, GPT-2 \citep{radford2019language}, with 124M parameters (small version), following prior ``controlled rearing'' work \citep{kallini2024mission, misra2024language, qin2024systematic}. Second, we train the hybrid LM, GPT-BERT, which uses masked next-token prediction (MNTP) combining autoregressive and masked objectives within a single Transformer architecture \citep{charpentier2024gpt}; GPT-BERT was found to be effective for small-scale training on the BabyLM challenge. Training details (including hyperparameters) can be found in Appendix \ref{app:train_compute_details}.

We investigated two sizes of each model, to investigate whether performance might vary at such small data scales, and in case of overfitting. We looked at GPT-2 small (124M), GPT-2 mini (39M), GPT-BERT base (119M), and GPT-BERT small (30M). For both model architectures, we train a separate tokenizer on the entire $BV_{Eng}$ data.\footnote{We do not train separate tokenizers for every individual experiment, as many individual $BV_{Eng}$ families contained too few unique types. As such, filtering our evaluation data using the intersection of the individual families' vocabulary would eliminate the vast majority of evaluation examples.} We train five separate seeds of each model for every experiment, reporting averaged results and seed variability. 

\subsection{Evaluation Benchmarks}

Drawing from the second BabyLM competition evaluation suite \citep{hu2024findings} and prior related work such as \citet{feng-etal-2024-child}, we selected benchmarks suitable to the scale and domain of our dataset, discussed below. Note that we filtered the evaluation data by our BabyView dataset's vocabulary. In particular, an evaluation example was kept only if every token in the example appeared at least once in $BV_{Eng}$. 

\textbf{Zorro} assesses the grammatical capabilities of SLMs via forced-choice judgments between minimal pairs of grammatical and ungrammatical sentences created from a limited vocabulary \citep{huebner2021babyberta}. We report average accuracy across individual Zorro tasks.

\textbf{Word Similarity} measures the ability of models to capture semantic similarities between pairs of words, allowing us to assess the semantic knowledge of our models \citep{zhuang2023visual}. We report Spearman correlations between human and model similarity judgments. More details in Appendix \ref{app:wordsim_details}. 

\textbf{COMPS} measures if models can demonstrate property inheritance, and infer that properties of superordinate concepts are inherited by subordinate concepts represented by nonce words \citep{misra-etal-2023-comps}. Similar to Zorro, models must choose between minimal pairs of sentences, and we report accuracy of assigning higher probability to the correct sentence.

\textbf{EWoK} evaluates models’ basic world knowledge by testing whether they can distinguish plausible from implausible scenarios given a context \citep{ewok}. The benchmark spans multiple domains of everyday knowledge, including social, physical, and spatial reasoning. We report accuracy of assigning higher probability to the correct continuation.



\subsection{Linguistic Feature Analysis}

To better understand why some BabyView family datasets produce stronger models than others, we compute a variety of linguistic features for each dataset (a full catalog of all 175 linguistic features is provided in Table \ref{app_tab:ling_feature_catalog_detailed} in Appendix~\ref{app:full_linguistic_features}). These features cover several categories of potential predictors of model performance:

\begin{tight_itemize}
\item \textbf{Lexical/distributional scale}: e.g., token and conversation counts, TTR/MATTR, entropy, Zipf/skewness, and related concentration statistics.
\item \textbf{Syntactic composition}: e.g., POS category proportions, POS bigram entropy/diversity, and dependency-parse structure metrics.
\item \textbf{Conversational/discourse structure}: e.g., turn-taking, question types, caregiver/child balance, overlap markers, repair/expansion cues.
\item \textbf{Mixture/composition metadata}: e.g., number of component families, age-distribution statistics, and cross-family divergence features.
\item \textbf{Semantic and predictability proxies}: e.g., adjacent-turn semantic similarity and normalized reference-LM perplexity features.
\item \textbf{Transcription/data-quality indicators}: e.g., unintelligible markers, partial-word and non-linguistic token rates.
\end{tight_itemize}


We analyze how well the linguistic features predict model performance across the 25 BabyView experiments discussed in \S\ref{subsec:exp_conditions} for the four different model-scale combinations discussed in \S\ref{subsec:model}. We focus on three evaluation targets: Zorro, WordSim, and COMPS.\footnote{We exclude EWoK due to poor scaling behavior in our BabyView-scale domain.} 
We run three complementary analyses for each model-target cell: (i) Spearman feature-target correlations, (ii) LassoCV with standardized predictors, and (iii) XGBoost regressors with feature-importance extraction. This provides a direct, method-agnostic view of which linguistic predictors recur across model-target settings.

\subsection{Predicting Children's Vocabulary Outcomes}

We were also interested in evaluating connections between model performance and children's learning outcomes \citep[see][]{portelance2023predicting}. One gold-standard measure for children's early vocabulary is the MacArthur-Bates Communicative Development Inventory \citep[CDI;][]{fenson2007macarthur, frank2021variability}, a validated parent-report vocabulary checklist. The families in BabyView filled out CDI forms approximately once every 3 months over the course of the study.

For each trained model, we computed the likelihood assigned to words appearing in the CDI vocabulary inventory. Specifically, we measured the average negative log-likelihood (NLL) of CDI words under each model by recursively splitting each example in our training corpora into 180-word contexts.\footnote{Required to fit them under the maximum sequence length of the corresponding models.} This metric serves as a proxy for how well the training data supports learning developmentally relevant vocabulary: lower NLL (i.e., higher probability) indicates that the model assigns higher probability to CDI vocabulary items. 

We calculated the age of acquisition (AoA) for each CDI item using the expressive vocabulary data in the BabyView CDIs, fitting a Bayesian binomial regression to the data predicting word production from age. We used weakly informative priors ($b_\textup{intercept} \sim \mathcal{N}(0, 2.5)$; $b_\textup{age} \sim \mathcal{N}(0.3, 0.1)$). We then calculated AoA as the negative of the intercept divided by the age slope; the AoA is the age at which 50\% of children are expected to know a word. We then compared a base frequency regression and an NLL regression: the base regression predicted AoA as a function of log frequency, concreteness, and their interactions with lexical category, while the NLL regression substituted model-derived NLL for log frequency -- theoretically a more sensitive measure of processing difficulty than log frequency \citep{portelance2023predicting}. To understand possible bias--variance tradeoffs from the training data, we considered three settings for obtaining NLL estimates: 1) averaging across the individual-family models, 2) the all-families models, and 3) the models trained on CHILDES (24.5M word English subset). We evaluated models in terms of their Akaike Information Criterion (AIC).

\section{Results}\label{sec:results}

\begin{figure}[t]
    \centering
    \includegraphics[width=\textwidth]{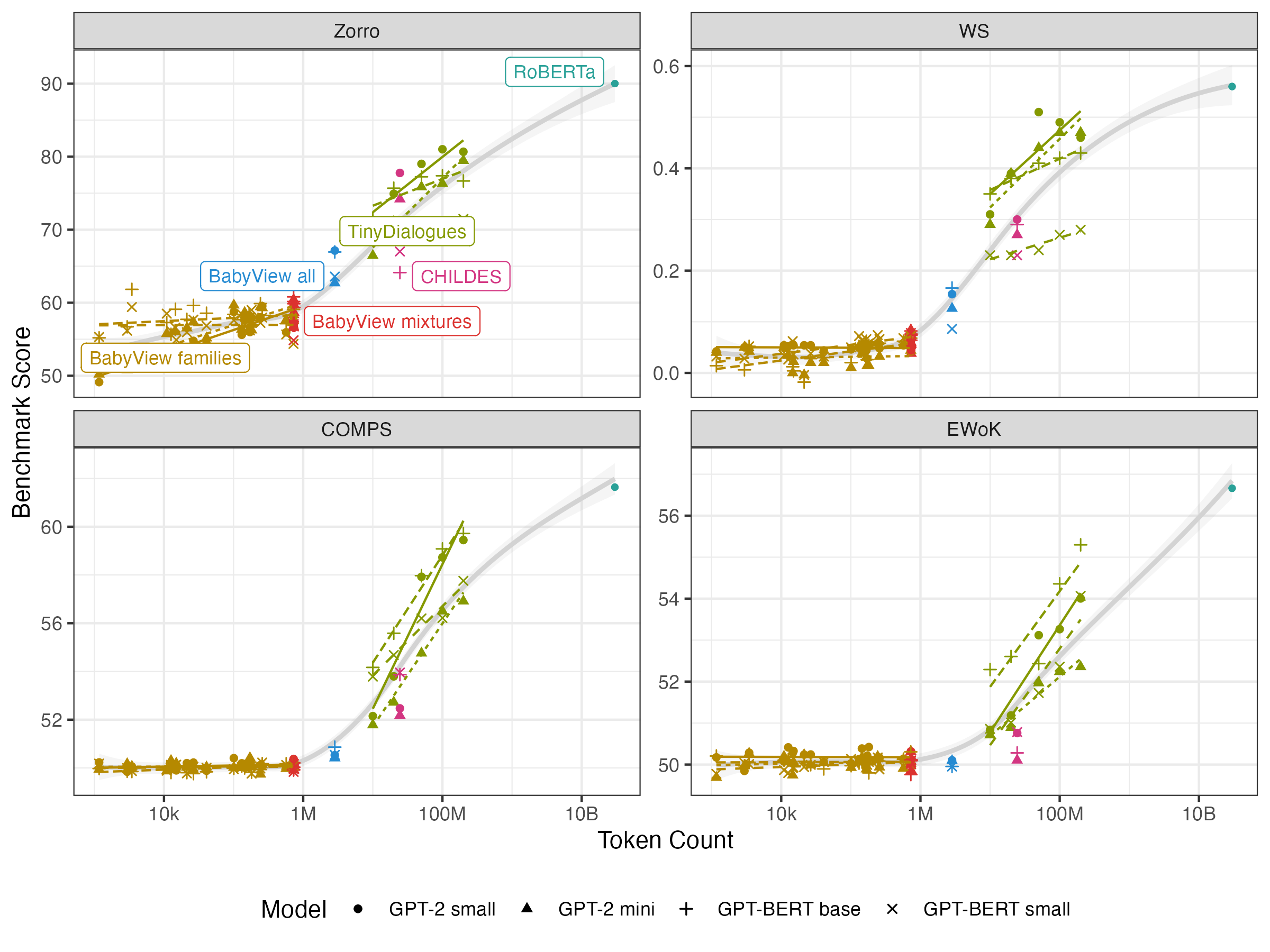}
    \caption{Broader performance scaling with data quantity across all four benchmarks. Each color denotes a separate dataset, with different markers showing different models. Gray curves show spline-smoothed average relationships across all data.} \label{fig:full_scaling_graphs}
\end{figure}

\begin{table*}[t]
\centering
\small
\resizebox{\textwidth}{!}{
\scalebox{0.90}{\begin{tabular}{llrrrrrrr}
\toprule
Model & Condition & \# of Fams & Tokens & TTR & Zorro & WordSim & COMPS & EWoK \\
\midrule
GPT-2 small & Mean of fams & 1 & 142,094 & -- & 55.19 $\pm$ 2.89 & 0.049 $\pm$ 0.007 & 50.09 $\pm$ 0.14 & 50.18 $\pm$ 0.15 \\
GPT-2 small & Top-1 & 1 & 725,600 & 0.026 & 58.09 $\pm$ 3.36 & 0.042 $\pm$ 0.025 & 50.19 $\pm$ 0.49 & 50.07 $\pm$ 0.36 \\
GPT-2 small & Top-2 & 2 & 727,986 & 0.025 & 59.58 $\pm$ 1.85 & 0.064 $\pm$ 0.022 & 50.37 $\pm$ 0.33 & 50.12 $\pm$ 0.13 \\
GPT-2 small & Top-5 & 5 & 727,883 & 0.026 & 56.70 $\pm$ 2.88 & 0.062 $\pm$ 0.008 & 50.22 $\pm$ 0.27 & 50.32 $\pm$ 0.05 \\
GPT-2 small & Top-10 & 10 & 735,309 & 0.027 & 57.14 $\pm$ 2.79 & 0.048 $\pm$ 0.020 & 50.22 $\pm$ 0.20 & 50.13 $\pm$ 0.21 \\
GPT-2 small & Top-20 & 20 & 734,234 & 0.026 & 56.59 $\pm$ 2.18 & 0.044 $\pm$ 0.015 & 50.03 $\pm$ 0.23 & 50.31 $\pm$ 0.29 \\
GPT-2 small & All fams & 20 & 2,841,873 & 0.013 & 67.14 $\pm$ 0.78 & 0.154 $\pm$ 0.018 & 50.53 $\pm$ 0.34 & 50.10 $\pm$ 0.44 \\
\midrule
GPT-2 mini & Mean of fams & 1 & 142,094 & -- & 55.92 $\pm$ 2.85 & 0.031 $\pm$ 0.018 & 50.06 $\pm$ 0.15 & 50.04 $\pm$ 0.16 \\
GPT-2 mini & Top-1 & 1 & 725,600 & 0.026 & 57.54 $\pm$ 1.98 & 0.044 $\pm$ 0.018 & 50.10 $\pm$ 0.16 & 50.03 $\pm$ 0.26 \\
GPT-2 mini & Top-2 & 2 & 727,986 & 0.025 & 60.42 $\pm$ 2.09 & 0.084 $\pm$ 0.040 & 50.14 $\pm$ 0.17 & 50.08 $\pm$ 0.19 \\
GPT-2 mini & Top-5 & 5 & 727,883 & 0.026 & 58.34 $\pm$ 1.20 & 0.060 $\pm$ 0.027 & 50.20 $\pm$ 0.36 & 50.31 $\pm$ 0.64 \\
GPT-2 mini & Top-10 & 10 & 735,309 & 0.027 & 59.74 $\pm$ 1.48 & 0.056 $\pm$ 0.026 & 50.13 $\pm$ 0.08 & 49.86 $\pm$ 0.09 \\
GPT-2 mini & Top-20 & 20 & 734,234 & 0.026 & 58.66 $\pm$ 1.68 & 0.038 $\pm$ 0.011 & 49.91 $\pm$ 0.19 & 49.85 $\pm$ 0.59 \\
GPT-2 mini & All fams & 20 & 2,841,873 & 0.013 & 62.71 $\pm$ 2.50 & 0.126 $\pm$ 0.025 & 50.41 $\pm$ 0.09 & 50.09 $\pm$ 0.25 \\
\midrule
GPT-BERT base & Mean of fams & 1 & 142,094 & -- & 57.77 $\pm$ 1.79 & 0.036 $\pm$ 0.024 & 50.00 $\pm$ 0.16 & 50.06 $\pm$ 0.11 \\
GPT-BERT base & Top-1 & 1 & 725,600 & 0.026 & 59.10 $\pm$ 1.58 & 0.068 $\pm$ 0.026 & 50.22 $\pm$ 0.27 & 50.31 $\pm$ 0.35 \\
GPT-BERT base & Top-2 & 2 & 727,986 & 0.025 & 60.17 $\pm$ 1.81 & 0.064 $\pm$ 0.011 & 49.93 $\pm$ 0.26 & 50.09 $\pm$ 0.17 \\
GPT-BERT base & Top-5 & 5 & 727,883 & 0.026 & 60.80 $\pm$ 2.80 & 0.082 $\pm$ 0.015 & 49.91 $\pm$ 0.21 & 49.76 $\pm$ 0.00 \\
GPT-BERT base & Top-10 & 10 & 735,309 & 0.027 & 59.35 $\pm$ 1.97 & 0.074 $\pm$ 0.005 & 50.21 $\pm$ 0.25 & 50.10 $\pm$ 0.15 \\
GPT-BERT base & Top-20 & 20 & 734,234 & 0.026 & 59.86 $\pm$ 2.45 & 0.076 $\pm$ 0.019 & 50.03 $\pm$ 0.42 & 50.12 $\pm$ 0.22 \\
GPT-BERT base & All fams & 20 & 2,841,873 & 0.013 & 66.95 $\pm$ 2.09 & 0.166 $\pm$ 0.005 & 50.86 $\pm$ 0.24 & 49.95 $\pm$ 0.44 \\
\midrule
GPT-BERT small & Mean of fams & 1 & 142,094 & -- & 56.94 $\pm$ 1.38 & 0.049 $\pm$ 0.021 & 50.03 $\pm$ 0.13 & 49.98 $\pm$ 0.10 \\
GPT-BERT small & Top-1 & 1 & 725,600 & 0.026 & 54.37 $\pm$ 1.63 & 0.072 $\pm$ 0.016 & 50.10 $\pm$ 0.10 & 50.00 $\pm$ 0.11 \\
GPT-BERT small & Top-2 & 2 & 727,986 & 0.025 & 54.86 $\pm$ 1.17 & 0.056 $\pm$ 0.021 & 49.84 $\pm$ 0.11 & 50.05 $\pm$ 0.15 \\
GPT-BERT small & Top-5 & 5 & 727,883 & 0.026 & 57.08 $\pm$ 1.74 & 0.078 $\pm$ 0.018 & 49.98 $\pm$ 0.25 & 50.05 $\pm$ 0.10 \\
GPT-BERT small & Top-10 & 10 & 735,309 & 0.027 & 56.88 $\pm$ 1.14 & 0.058 $\pm$ 0.019 & 50.16 $\pm$ 0.34 & 50.01 $\pm$ 0.20 \\
GPT-BERT small & Top-20 & 20 & 734,234 & 0.026 & 56.76 $\pm$ 1.08 & 0.062 $\pm$ 0.015 & 49.96 $\pm$ 0.15 & 50.10 $\pm$ 0.36 \\
GPT-BERT small & All fams & 20 & 2,841,873 & 0.013 & 63.58 $\pm$ 1.73 & 0.086 $\pm$ 0.015 & 50.54 $\pm$ 0.49 & 49.95 $\pm$ 0.18 \\
\bottomrule
\end{tabular}}}
\caption{Major BabyView results (averaged across 5 seeds per experiment). \textit{Mean of fams} reports mean and standard deviation across the 20 individual-family datasets. Pooled rows (Top-2/5/10/20) are proportional mixtures with seed-level variability. \textit{Top-1} is the largest single family; \textit{All fams} is the all-families model trained on the concatenated corpus. Zorro, COMPS, and EWoK are percentages (higher is better); WordSim is Spearman correlation (higher is better). TTR is type–token ratio (unique word types divided by total tokens).}\label{tab:BV_diversity_summary}
\end{table*}

\begin{figure}[t]
    \centering
    \includegraphics[width=\textwidth]{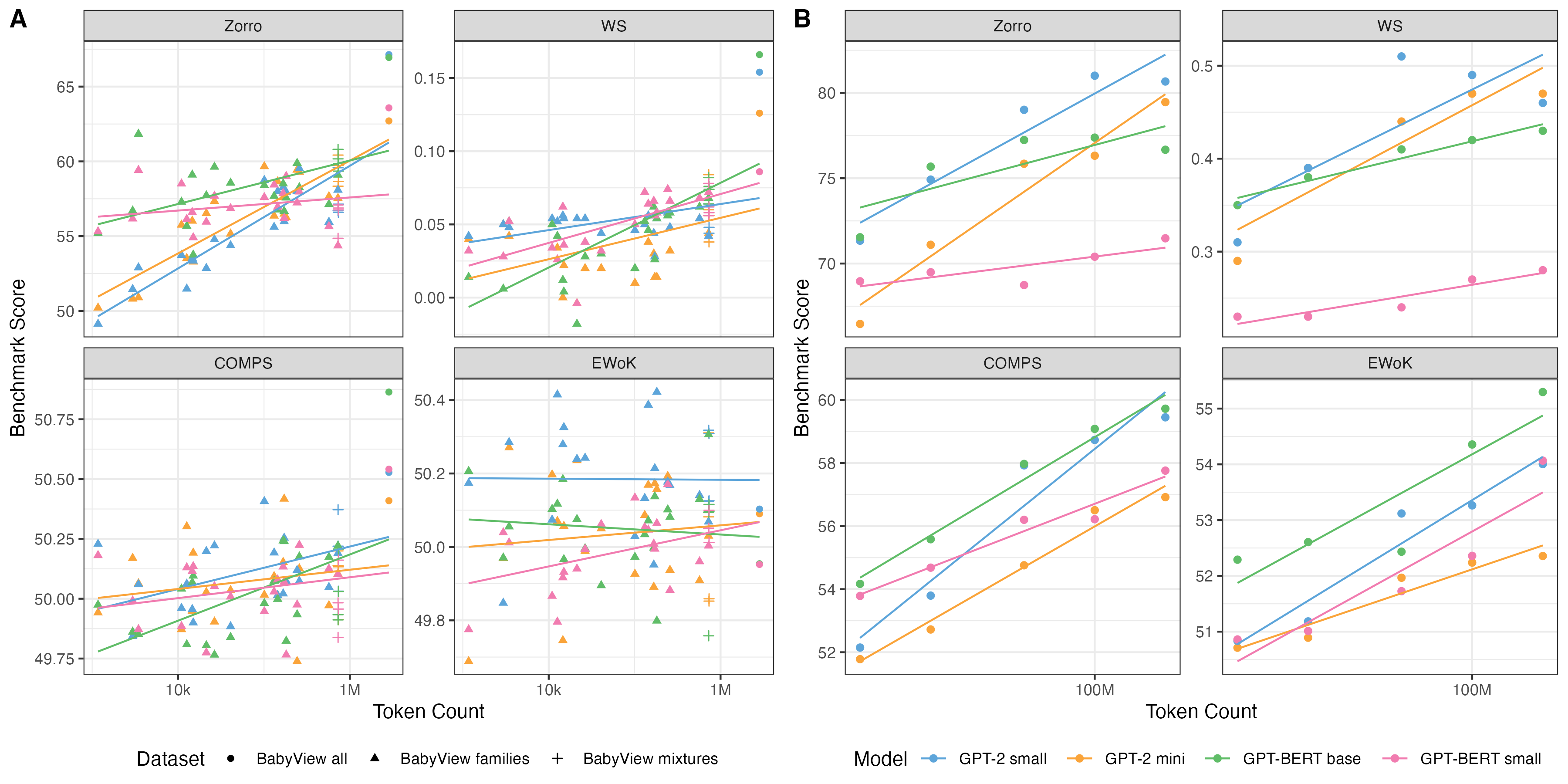}
    \caption{Performance scaling of (A) BabyView experiments and (B) TinyDialogues experiments across all four benchmarks. Each color denotes a separate model; different markers show different experimental conditions. Lines of best fit show average linear relationships.} \label{fig:combined}
\end{figure}

\subsection{Scaling at Child-Scale Data Regimes}

\paragraph{Scaling within BabyView datasets.}

We first examine scaling behavior within BabyView itself. Each family dataset contains a different number of tokens, allowing us to study how model performance varies as a function of input size. Major results are shown in Table \ref{tab:BV_diversity_summary} and Figures~\ref{fig:full_scaling_graphs} and \ref{fig:combined}A.\footnote{Explicit individual family results are in Tables \ref{tab:ind_family_results_gpt2_small} to \ref{tab:ind_family_results_gptbert_small} in Appendix \ref{app:ind_family_results}.} Across benchmarks other than EWoK (world knowledge) which shows little to no signal, we observe a positive relationship between dataset size and model performance, but scaling slopes are strong for grammatical knowledge (Zorro) and weaker for WordSim and COMPS, which represent semantic knowledge. Larger datasets generally produce stronger models, although results were notably variable between families. 

\paragraph{Mixture vs single-family datasets.}

We next compared models trained on mixtures of families with those trained on individual families (Table \ref{tab:BV_diversity_summary}). More families were not always monotonically better: at matched token budgets, some pooled mixtures outperform the largest individual family (Top-1), but the pattern was benchmark and architecture dependent. Importantly, adding families did not increase raw lexical diversity as measured by type–token ratio (TTR). This suggests that the benefits of pooling, when they occur, may arise less from adding many new word types and more from increased contextual, discourse, speaker, and/or interactional diversity. See \S\ref{subsec:perf_var_ling_analysis} below for further analysis of these features.

\paragraph{Broader scaling behavior.} 
To extend our scaling curves, we trained our models on increasing amounts of the synthetic TinyDialogues (TD) data ranging from 10M to 200M tokens (Figure~\ref{fig:combined}B). Across all four benchmarks, model performance improves $\sim$ linearly with the logarithm of training tokens, consistent with classical scaling law behavior. We further compare to results on the 24.5M word English-subset of CHILDES, following \citet{feng-etal-2024-child}, and a pretrained RoBERTa-base on 30B tokens (serving as a topline). As seen in Figure~\ref{fig:full_scaling_graphs}, all four benchmarks exhibit classical scaling trends, and scaling is notably different than with BabyView data. To quantify this trend, we fit linear models to each model and benchmark predicting performance as a function of dataset, log(tokens), and their interaction. Of the 16 models, all but four (GPT-BERT scaling on Zorro and WordSim) showed significant interactions, indicating different scaling curves for BabyView vs. TinyDialogues. One possibility is that the synthetic TinyDialogues data is richer and more variable in content than true child data, leading to stronger scaling. 

\subsection{Performance Variability Across Families}\label{subsec:perf_var_ling_analysis}

Despite similar dataset sizes, models trained on different families showed noticeably different performance, as seen in Figure~\ref{fig:combined}A. 
Although there was a positive relationship between token count and performance, families with comparable sizes produced models whose scores differed by several points, as did different mixtures of families. Relative to seed-level variability, between-experiment spread is comparable and can be larger for specific model-metric pairs. 
Thus, dataset size alone likely does not fully determine learning outcomes. 

\begin{table}[t]
\centering
\small
\setlength{\tabcolsep}{4pt}
\renewcommand{\arraystretch}{1.05}

\newcommand{\featureline}{\specialrule{0.15pt}{0pt}{0pt}}

\resizebox{\textwidth}{!}{\begin{tabular}{lccccccccc}
\toprule
\textbf{Feature} & \textbf{Top10} & \textbf{Meth.} & \textbf{Spear.} & \textbf{Lasso} & \textbf{XGB} & \makecell{\textbf{Mean}\\\textbf{rank}} & \makecell{\textbf{Mean}\\\textbf{$|\rho|$}} & \makecell{\textbf{Mean}\\\textbf{$|\beta|$}} & \makecell{\textbf{Mean}\\\textbf{imp.}} \\
\midrule

Bigram mutual information 
& 9 & 3 & 3 & 1 & 5 & 5.56 & 0.72 & 0.153 & 0.060 \\

Caregiver POS-tagged token count 
& 7 & 3 & 4 & 1 & 2 & 4.29 & 0.74 & 0.016 & 0.062 \\

Child POS-tagged token count
& 7 & 3 & 1 & 4 & 2 & 2.00 & 0.86 & 0.604 & 0.356 \\

Child-to-caregiver semantic pair count 
& 6 & 3 & 2 & 2 & 2 & 6.17 & 0.80 & 0.005 & 0.050 \\

POS bigram entropy 
& 6 & 3 & 1 & 1 & 4 & 3.33 & 0.56 & 0.263 & 0.069 \\

Mean KL divergence from other datasets 
& 5 & 3 & 3 & 1 & 1 & 3.60 & 0.71 & 0.500 & 0.053 \\

Total parse-eligible utterances 
& 5 & 3 & 1 & 1 & 3 & 3.40 & 0.85 & 0.012 & 0.149 \\

Hapax ratio 
& 5 & 3 & 1 & 1 & 3 & 6.60 & 0.48 & 0.264 & 0.067 \\

Trigram entropy 
& 5 & 2 & 4 & 0 & 1 & 5.00 & 0.74 & 0.000 & 0.075 \\

Number of conversations 
& 5 & 1 & 0 & 5 & 0 & 2.00 & 0.00 & 0.178 & 0.000 \\

\bottomrule
\end{tabular}}

\caption{
Top linguistic predictors across statistical analyses, identified and sorted by Top10 hit. 
\textbf{Top10} counts the number of times a feature appears in top-10 lists across all methods and experimental conditions. 
\textbf{Meth.} is the number of distinct methods (Spearman, Lasso, XGBoost) that select the feature at least once. 
\textbf{Spear.}, \textbf{Lasso}, and \textbf{XGB} report method-specific top-10 counts. The following mean columns compute average across top-10 appearances. 
\textbf{Mean rank} is the average within-list rank when selected. 
\textbf{Mean $|\rho|$} is the average absolute Spearman correlation. 
\textbf{Mean $|\beta|$} is the average absolute lasso coefficient (linear effect size under regularization). 
\textbf{Mean imp.} is the average XGBoost feature importance (nonlinear contribution to predictive performance). Full results (all linguistic features) in Table \ref{app_tab:ling_full_grouped}. 
}
\label{tab:linguistic_predictors}
\end{table}



Table~\ref{tab:linguistic_predictors} summarizes the top linguistic features found to be most influential on downstream performance across the 25 BabyView experiments and 4 model-size combinations (see also Table \ref{app_tab:predictive_model_metrics} in Appendix \ref{app:full_linguistic_features}).
The strongest predictors included distributional and syntactic features such as bigram mutual information, POS bigram entropy, and POS-tagged token coverage (caregiver and child). Further, divergence and coverage features (e.g., semantic pair coverage, mean KL from other datasets, parse-eligible utterance count, and number of conversations) were also related to downstream performance, suggesting that models exposed to broader and more consistently structured linguistic contexts learn more effectively. These variables indicate that multiple facets of input quality, composition, and coverage shape model performance, rather than dataset size alone.



\begin{figure}
    \centering
    \includegraphics[width=\textwidth]{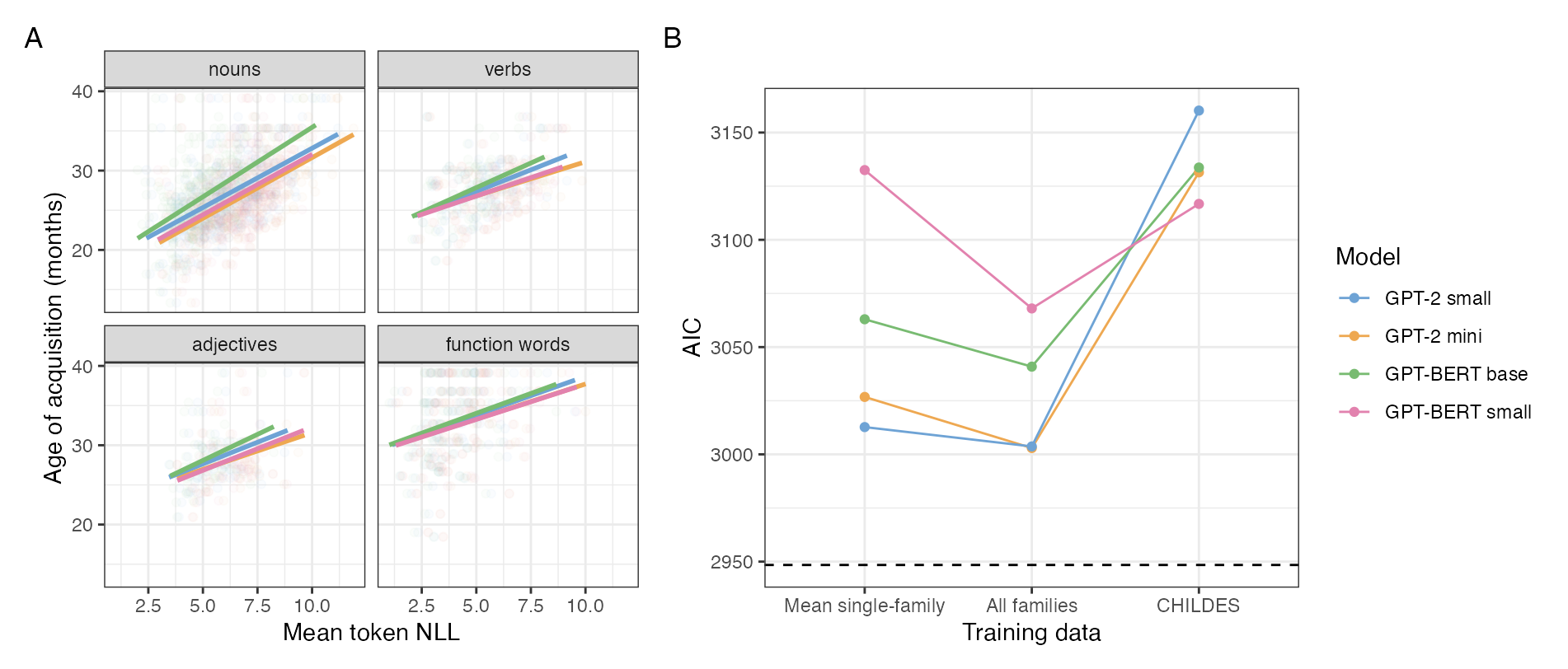}
    \caption{(A) Relationship between mean token NLL and AoA for CDI words by lexical category, for models trained on data from all families. (B) AIC of the regressions predicting AoA from mean token NLL. Dashed line indicates AIC of the base frequency model.}
    \label{fig:cdi}
\end{figure}

\subsection{Connections to Child Vocabulary Outcomes}\label{subsec:CDI_results}

We examined the relationship between model token likelihood and children's reported vocabulary outcomes on the CDI, using the age of acquisition of CDI words. 
While token NLL appeared to show some relationships with AoA (Figure~\ref{fig:cdi}A for models trained on data from all families), none of the regressions containing model NLL values attained a lower AIC than the base frequency model (Figure~\ref{fig:cdi}B).
We note that there are moderate negative correlations between log frequency and mean token NLL ($r \in [-.56, -.42]$), suggesting that NLL values share some variance with log frequency but may be capturing additional noise.

\section{Discussion} 

We examined the scaling properties of SLMs trained on data from children's linguistic experiences, and -- inspired by children's language learning -- asked whether features of particular families' language environment lead to better scaling. Classical scaling laws appear to persist even in extremely small data regimes for grammaticality: model performance improved linearly in the logarithm of training tokens. Models scaled less well on semantic tasks, however, 
and there was almost no increase in performance on world knowledge for our BabyView experiments. Perhaps the restricted data from interactions in children's home environments is insufficient input for learning about the broader world. More diverse input data -- e.g., from books or experiences outside the home -- might play an important role for semantic generalization and world knowledge acquisition. 

Variability across individual language environments was substantial. Models trained on different families’ input achieved markedly different performance levels even when dataset sizes were similar. This finding parallels observations in language development research showing that children experience highly variable language environments, and that these environments relate to their learning outcomes \citep{hart1997meaningful, weisleder2013talking, coffey2025strong}. 
Our linguistic feature analysis suggests that multiple factors, including distributional structure, syntactic composition and coverage, and semantic coverage, play an important role in shaping model performance across multiple tasks, aligning with prior work showing that modifying the structure and composition of input representations can significantly affect LM generation quality \citep[e.g.,][]{feng-etal-2021-sapphire}. 
This again mirrors findings from child language development that emphasize quality over pure quantity of input \citep{anderson2021linking, masek2021beyond,rowe2012longitudinal, rowe2020analyzing}. This also aligns with our above finding, suggesting that more diverse and high-quality input data is likely necessary for effective LM learning, especially at such small scales.

Finally, the relationship between model CDI likelihood and children’s vocabulary outcomes suggests that some properties of child-directed input may influence both machine and human language learning. While these results are preliminary, they point toward the possibility of using LMs as computational probes for studying the structure of individual children's linguistic environments, much as has been done with larger scale data already \citep[e.g.,][]{portelance2023predicting}. 

\section{Conclusion}

Together, these findings highlight the importance of considering not only the quantity of language input but also its composition, structure, and coverage.
We have only scratched the surface of this goal here: future work should move beyond transcribed speech to consider the grounded, multimodal nature of language exposure. 
Further, from a psychological perspective, the generalizability of our work here is limited by the relatively small sample size we examined; while the BabyView corpus is the largest of its kind, it still represents only one language (English) and a small, selected subset of families. 
Nevertheless, we hope that our work sets the stage for future investigations of children's language learning environments using language models. 
Understanding what makes language data efficient for learning may ultimately inform both the design of more data-efficient language models and scientific theories of human language acquisition. 

\section*{Acknowledgments}
We gratefully acknowledge Modal for providing a portion of the compute resources that enabled this work.\\






\bibliography{colm2026_conference}
\bibliographystyle{colm2026_conference}

\appendix

\newcommand*{\escape}[1]{\texttt{\textbackslash#1}}
\newcommand*{\escapeI}[1]{\texttt{\expandafter\string\csname #1\endcsname}}
\newcommand*{\escapeII}[1]{\texttt{\char`\\#1}}

\newpage
\section{Dataset Examples \& Preprocessing Details}\label{app:dataset_details}

Following \citet{feng-etal-2024-child}, we preprocess training examples in both BabyView and TinyDialogues in the same way. Double asterisks surround speaker labels, double newline tokens separate utterances, and an end-of-text token marks the end of the conversation. This format was consistent across all conversations in both BabyView and TinyDialogues.

\paragraph{BabyView example:} \textit{**MOT**: Put it in the oven for baby and me. \escape{n}\escape{n} **MOT**: Oh, hold on. \escape{n}\escape{n} **MOT**: Let's try to put it on. \escape{n}\escape{n} **CHI**: Whoopsie. \escape{n}\escape{n} **MOT**: Whoopsie. \escape{n}\escape{n} **OCHI**: Hey, Rosa, look at me. \escape{n}\escape{n} **MOT**: Look at me. \escape{n}\escape{n} (...) \texttt{<|endoftext|>}}

\paragraph{TinyDialogues example:} \textit{**Teacher**: Alright, everyone, it's time to clean up! **Child**, can you please help me by putting the crayons back in the box? \escape{n}\escape{n} **Child**: Yes! I can do that. The box is empty so I'll fill it up! \escape{n}\escape{n} **Teacher**: Thank you, that's very helpful. Make sure the lids are on tight so they don't dry out. \escape{n}\escape{n} **Child**: I did it! Look, they're all inside now. \escape{n}\escape{n} (...) \texttt{<|endoftext|>}}

Training data was set up so that each line corresponded to a single conversation. Removing the child's speech from the dialogue would create incoherent training data that lacked context; thus, consistent with previous work leveraging transcripts of children's language environment, we include both speaker labels and the child's own speech in our training data \citep{huebner2021babyberta, feng-etal-2024-child} . 

\section{Further Training \& Compute Details}\label{app:train_compute_details}
\label{appendix:compute_details}

We trained a separate tokenizer for each model on $BV_{Eng}$, TinyDialogues, and CHILDES. We use a vocab size of 50k for GPT-2, 16k for GPT-BERT base, and 8k for GPT-BERT small. For each training run, the best checkpoint was selected based on lowest validation loss.

We pretrain GPT-2 from scratch, using a linear LR scheduler with no warmup, varying batch sizes per GPU (4 to 64 depending on the GPU and experiment), and Adam optimizer with $\beta=(0.9,0.999)$ and $\epsilon=1e-08$. During training, GPT-2 processes data in 1024-token chunks. To allow the models to distinguish between individual conversations and training examples, we add an \textit{end-of-text} token to the end of each conversation. We train GPT-2 small for up to 20 epochs for all experiments, and GPT-2 mini for up to 50 epochs for BabyView experiments and 20 epochs for TinyDialogues and CHILDES. Our GPT-2 experiments are based on HuggingFace's causal LM training code, with modifications for our experiments.

We pretrain GPT-BERT from scratch, using a cosine LR scheduler with 0.1 weight decay. On the BabyView experiments, we use a sequence length which increases throughout training (128 to 256 after 70\% of training and 512 after 90\% of training) while the batch size decreases accordingly with each sequence length increase (down to 1/2 and 1/4 -- from 2048 for small and 1024 for base down to 512 and 256, respectively). For our TinyDialogues and CHILDES training, we keep sequence length (128) and batch size (512 for small, 256 for base) fixed throughout training.

Further, due to GPT-BERT's hybrid objective that mixes autoregressive next-token prediction and masked-token prediction, we try different causal:mask ratios (e.g., 1:1 means half causal and half masked training workers), while the implementation maps this to a masked-worker fraction under GPU divisibility constraints. We search through \{0:1, 1:7, 1:3, 1:1, 3:1, 7:1, 1:0\} ratios, and settle on 1:1 for our final training experiments due to the most stable evaluation and scaling behavior. Our GPT-BERT experiments are based on the original authors' GitHub repo and code\footnote{\url{https://github.com/ltgoslo/gpt-bert}}, with modifications for our experiments.

We train 5 seeds $\{42,0,123,321,9000\}$ of each model for every $BV_{Eng}$ experiment, and a single seed (either 42 or 0) for TinyDialogues and CHILDES. We searched through different values of the learning rate (LR) for GPT-2 training. Specifically, $LR = \{1e-06,5e-06,1e-05,5e-05,1e-04,3e-04,5e-04,1e-03\}$. Through initial experiments, we found that $LR = 1e-04$ led to the best convergence behavior for small and $LR = 3e-04$ for mini, and used that for all our training experiments. We do the same for GPT-BERT, searching through $LR = \{3e-5, 5e-5, 1e-4, 5e-4, 1e-3, 3e-3, 5e-3, 1e-2, 3e-2, 5e-2\}$, and choose $LR = 1e-02$ for both model sizes for BabyView experiments, and $LR = 5e-3$ for base and $LR = 1e-2$ for small for TinyDialogues and CHILDES.

Due to GPT-BERT's architecture, we report three ways (inference modes) of conducting Zorro evaluation of GPT-BERT: causal, bidirectional, and fused. We find that results can vary depending on the inference mode chosen and the causal:mask ratio used for training. Causal applies an autoregressive triangular attention mask, bidirectional uses full attention, and fused computes both passes and sums their logits before scoring. For each set of experiments, we report the Zorro inference mode that results in the most stable evaluation and scaling behavior.

Our experiments were run on varying GPUs. This included a single RTX 3090TI and up to eight A40s, A100s, H100s, and H200s. Training time varied by the type and number of GPUs used and the particular experiment.

\section{Word Similarity Benchmark Details}\label{app:wordsim_details}
Following \citet{zhuang2023visual}, we extract word embeddings from the hidden layers of each model and compute pairwise cosine similarities. The best layer of each model is chosen. We average results across several word similarity benchmarks including RG-65 \citep{rg-65}, WordSim-353 \citep{wordsim-353}, SimLex-999 \citep{hill-etal-2015-simlex}, SimVerb-3500 \citep{gerz-etal-2016-simverb}, and MEN (MTest-3000) \citep{bruni-etal-2012-distributional}.

\begin{table}[t]
\centering
\resizebox{\textwidth}{!}{\begin{tabular}{lrrrrrr}
\toprule
Family & Age (months) & Tokens & Zorro & WordSim & COMPS & EWoK \\
\midrule
BV-fam-S00400001 & 31 & 725,600 & 58.09 $\pm$ 3.36 & 0.042 $\pm$ 0.025 & 50.19 $\pm$ 0.49 & 50.07 $\pm$ 0.36 \\
BV-fam-S00510002 & 11 & 569,788 & 55.95 $\pm$ 2.74 & 0.054 $\pm$ 0.015 & 50.05 $\pm$ 0.42 & 50.14 $\pm$ 0.20 \\
BV-fam-S00360001 & 8 & 257,817 & 59.53 $\pm$ 1.83 & 0.048 $\pm$ 0.022 & 50.07 $\pm$ 0.15 & 50.17 $\pm$ 0.30 \\
BV-fam-S00220001 & 5 & 242,954 & 58.02 $\pm$ 2.19 & 0.058 $\pm$ 0.029 & 50.12 $\pm$ 0.34 & 50.18 $\pm$ 0.36 \\
BV-fam-S00400002 & 28 & 181,317 & 57.95 $\pm$ 2.19 & 0.054 $\pm$ 0.031 & 50.07 $\pm$ 0.35 & 50.42 $\pm$ 0.80 \\
BV-fam-S00320002 & 7 & 171,656 & 55.98 $\pm$ 2.31 & 0.028 $\pm$ 0.024 & 50.25 $\pm$ 0.32 & 50.21 $\pm$ 0.27 \\
BV-fam-S00370001 & 8 & 167,136 & 58.28 $\pm$ 1.92 & 0.044 $\pm$ 0.023 & 50.02 $\pm$ 0.27 & 49.95 $\pm$ 0.20 \\
BV-fam-S00430001 & 31 & 144,253 & 58.00 $\pm$ 2.25 & 0.056 $\pm$ 0.030 & 50.01 $\pm$ 0.24 & 50.39 $\pm$ 0.89 \\
BV-fam-S00510001 & 11 & 131,120 & 55.61 $\pm$ 3.61 & 0.050 $\pm$ 0.023 & 50.19 $\pm$ 0.19 & 50.13 $\pm$ 0.31 \\
BV-fam-S00720001 & 38 & 100,807 & 58.74 $\pm$ 1.71 & 0.046 $\pm$ 0.049 & 50.41 $\pm$ 0.41 & 50.03 $\pm$ 0.35 \\
BV-fam-S00490001 & 27 & 40,835 & 54.38 $\pm$ 5.82 & 0.044 $\pm$ 0.032 & 49.88 $\pm$ 0.23 & 50.06 $\pm$ 0.24 \\
BV-fam-S00230001 & 22 & 26,501 & 54.78 $\pm$ 4.63 & 0.054 $\pm$ 0.030 & 50.22 $\pm$ 0.16 & 50.24 $\pm$ 0.30 \\
BV-fam-S00440001 & 26 & 21,313 & 52.86 $\pm$ 6.37 & 0.054 $\pm$ 0.017 & 50.20 $\pm$ 0.20 & 50.24 $\pm$ 0.17 \\
BV-fam-S00360002 & 8 & 15,006 & 53.35 $\pm$ 2.22 & 0.054 $\pm$ 0.021 & 49.90 $\pm$ 0.51 & 50.33 $\pm$ 0.57 \\
BV-fam-S00460001 & 11 & 14,618 & 53.52 $\pm$ 4.27 & 0.056 $\pm$ 0.021 & 49.96 $\pm$ 0.41 & 50.28 $\pm$ 0.12 \\
BV-fam-S00350002 & 8 & 12,614 & 51.48 $\pm$ 2.89 & 0.052 $\pm$ 0.024 & 50.06 $\pm$ 0.53 & 50.41 $\pm$ 0.05 \\
BV-fam-S00550001 & 31 & 10,952 & 53.73 $\pm$ 3.18 & 0.054 $\pm$ 0.023 & 49.96 $\pm$ 0.53 & 50.07 $\pm$ 0.42 \\
BV-fam-S00340002 & 7 & 3,457 & 52.90 $\pm$ 2.86 & 0.048 $\pm$ 0.028 & 50.06 $\pm$ 0.20 & 50.28 $\pm$ 0.29 \\
BV-fam-S00350001 & 25 & 2,959 & 51.45 $\pm$ 4.72 & 0.050 $\pm$ 0.014 & 49.84 $\pm$ 0.32 & 49.85 $\pm$ 0.35 \\
BV-fam-S01010001 & 23 & 1,170 & 49.13 $\pm$ 5.16 & 0.042 $\pm$ 0.016 & 50.23 $\pm$ 0.48 & 50.17 $\pm$ 0.50 \\
\bottomrule
\end{tabular}}
\caption{Individual-family BabyView results for GPT-2 small with seed-level variability, sorted by training tokens. Age is months at onboarding time of that family's given child.}\label{tab:ind_family_results_gpt2_small}
\end{table}

\begin{table}[t]
\centering
\resizebox{\textwidth}{!}{\begin{tabular}{lrrrrrr}
\toprule
Family & Age (months) & Tokens & Zorro & WordSim & COMPS & EWoK \\
\midrule
BV-fam-S00400001 & 31 & 725,600 & 57.54 $\pm$ 1.98 & 0.044 $\pm$ 0.018 & 50.10 $\pm$ 0.16 & 50.03 $\pm$ 0.26 \\
BV-fam-S00510002 & 11 & 569,788 & 57.65 $\pm$ 2.52 & 0.054 $\pm$ 0.027 & 49.97 $\pm$ 0.20 & 49.91 $\pm$ 0.10 \\
BV-fam-S00360001 & 8 & 257,817 & 59.29 $\pm$ 0.95 & 0.032 $\pm$ 0.033 & 50.13 $\pm$ 0.29 & 49.94 $\pm$ 0.36 \\
BV-fam-S00220001 & 5 & 242,954 & 59.43 $\pm$ 1.08 & 0.058 $\pm$ 0.018 & 49.74 $\pm$ 0.45 & 50.19 $\pm$ 0.30 \\
BV-fam-S00400002 & 28 & 181,317 & 56.25 $\pm$ 3.03 & 0.014 $\pm$ 0.024 & 50.07 $\pm$ 0.32 & 50.16 $\pm$ 0.25 \\
BV-fam-S00320002 & 7 & 171,656 & 56.49 $\pm$ 2.74 & 0.014 $\pm$ 0.019 & 50.42 $\pm$ 0.27 & 50.17 $\pm$ 0.38 \\
BV-fam-S00370001 & 8 & 167,136 & 57.59 $\pm$ 1.09 & 0.030 $\pm$ 0.019 & 50.15 $\pm$ 0.26 & 49.89 $\pm$ 0.12 \\
BV-fam-S00430001 & 31 & 144,253 & 58.65 $\pm$ 2.60 & 0.038 $\pm$ 0.022 & 50.01 $\pm$ 0.24 & 50.17 $\pm$ 0.31 \\
BV-fam-S00510001 & 11 & 131,120 & 56.36 $\pm$ 1.33 & 0.052 $\pm$ 0.025 & 50.09 $\pm$ 0.25 & 50.09 $\pm$ 0.36 \\
BV-fam-S00720001 & 38 & 100,807 & 59.66 $\pm$ 1.68 & 0.010 $\pm$ 0.022 & 50.02 $\pm$ 0.08 & 49.93 $\pm$ 0.66 \\
BV-fam-S00490001 & 27 & 40,835 & 55.16 $\pm$ 3.38 & 0.020 $\pm$ 0.035 & 50.04 $\pm$ 0.38 & 50.05 $\pm$ 0.21 \\
BV-fam-S00230001 & 22 & 26,501 & 57.34 $\pm$ 2.53 & 0.020 $\pm$ 0.027 & 49.90 $\pm$ 0.49 & 49.99 $\pm$ 0.15 \\
BV-fam-S00440001 & 26 & 21,313 & 56.53 $\pm$ 3.06 & -0.004 $\pm$ 0.024 & 50.03 $\pm$ 0.29 & 50.24 $\pm$ 0.27 \\
BV-fam-S00360002 & 8 & 15,006 & 53.29 $\pm$ 2.73 & 0.022 $\pm$ 0.022 & 50.19 $\pm$ 0.44 & 50.06 $\pm$ 0.42 \\
BV-fam-S00460001 & 11 & 14,618 & 56.02 $\pm$ 4.22 & 0.000 $\pm$ 0.029 & 49.95 $\pm$ 0.34 & 49.75 $\pm$ 0.43 \\
BV-fam-S00350002 & 8 & 12,614 & 53.52 $\pm$ 2.64 & 0.034 $\pm$ 0.041 & 50.30 $\pm$ 0.35 & 50.07 $\pm$ 0.13 \\
BV-fam-S00550001 & 31 & 10,952 & 55.75 $\pm$ 2.61 & 0.050 $\pm$ 0.025 & 49.87 $\pm$ 0.33 & 50.20 $\pm$ 0.25 \\
BV-fam-S00340002 & 7 & 3,457 & 50.90 $\pm$ 3.88 & 0.042 $\pm$ 0.029 & 50.06 $\pm$ 0.36 & 50.27 $\pm$ 0.25 \\
BV-fam-S00350001 & 25 & 2,959 & 50.81 $\pm$ 1.67 & 0.050 $\pm$ 0.017 & 50.17 $\pm$ 0.32 & 49.97 $\pm$ 0.58 \\
BV-fam-S01010001 & 23 & 1,170 & 50.21 $\pm$ 4.74 & 0.040 $\pm$ 0.024 & 49.94 $\pm$ 0.30 & 49.69 $\pm$ 0.50 \\
\bottomrule
\end{tabular}}
\caption{Individual-family BabyView results for GPT-2 mini with seed-level variability, sorted by training tokens. Age is months at onboarding time of that family's given child.}\label{tab:ind_family_results_gpt2_mini}
\end{table}

\begin{table}[t]
\centering
\resizebox{\textwidth}{!}{\begin{tabular}{lrrrrrr}
\toprule
Family & Age (months) & Tokens & Zorro & WordSim & COMPS & EWoK \\
\midrule
BV-fam-S00400001 & 31 & 725,600 & 59.10 $\pm$ 1.58 & 0.068 $\pm$ 0.026 & 50.22 $\pm$ 0.27 & 50.31 $\pm$ 0.35 \\
BV-fam-S00510002 & 11 & 569,788 & 57.13 $\pm$ 1.69 & 0.062 $\pm$ 0.013 & 50.17 $\pm$ 0.20 & 50.13 $\pm$ 0.17 \\
BV-fam-S00360001 & 8 & 257,817 & 58.26 $\pm$ 0.85 & 0.058 $\pm$ 0.015 & 50.18 $\pm$ 0.29 & 50.08 $\pm$ 0.27 \\
BV-fam-S00220001 & 5 & 242,954 & 59.88 $\pm$ 1.82 & 0.056 $\pm$ 0.009 & 49.93 $\pm$ 0.23 & 50.10 $\pm$ 0.17 \\
BV-fam-S00400002 & 28 & 181,317 & 57.57 $\pm$ 1.58 & 0.054 $\pm$ 0.015 & 49.82 $\pm$ 0.28 & 49.80 $\pm$ 0.27 \\
BV-fam-S00320002 & 7 & 171,656 & 56.68 $\pm$ 0.94 & 0.026 $\pm$ 0.011 & 50.24 $\pm$ 0.34 & 50.14 $\pm$ 0.35 \\
BV-fam-S00370001 & 8 & 167,136 & 58.51 $\pm$ 1.41 & 0.062 $\pm$ 0.022 & 50.24 $\pm$ 0.26 & 50.00 $\pm$ 0.30 \\
BV-fam-S00430001 & 31 & 144,253 & 56.77 $\pm$ 1.13 & 0.046 $\pm$ 0.021 & 50.00 $\pm$ 0.26 & 50.07 $\pm$ 0.09 \\
BV-fam-S00510001 & 11 & 131,120 & 57.68 $\pm$ 2.73 & 0.052 $\pm$ 0.031 & 50.07 $\pm$ 0.22 & 50.03 $\pm$ 0.45 \\
BV-fam-S00720001 & 38 & 100,807 & 58.40 $\pm$ 1.95 & 0.020 $\pm$ 0.016 & 49.98 $\pm$ 0.17 & 49.97 $\pm$ 0.18 \\
BV-fam-S00490001 & 27 & 40,835 & 58.56 $\pm$ 1.69 & 0.030 $\pm$ 0.019 & 49.84 $\pm$ 0.04 & 49.90 $\pm$ 0.25 \\
BV-fam-S00230001 & 22 & 26,501 & 59.63 $\pm$ 1.46 & 0.028 $\pm$ 0.018 & 49.77 $\pm$ 0.20 & 50.00 $\pm$ 0.32 \\
BV-fam-S00440001 & 26 & 21,313 & 57.72 $\pm$ 1.90 & -0.018 $\pm$ 0.008 & 49.80 $\pm$ 0.38 & 50.08 $\pm$ 0.34 \\
BV-fam-S00360002 & 8 & 15,006 & 53.75 $\pm$ 1.85 & 0.004 $\pm$ 0.011 & 50.10 $\pm$ 0.27 & 49.97 $\pm$ 0.30 \\
BV-fam-S00460001 & 11 & 14,618 & 59.10 $\pm$ 1.34 & 0.012 $\pm$ 0.008 & 50.07 $\pm$ 0.32 & 50.18 $\pm$ 0.28 \\
BV-fam-S00350002 & 8 & 12,614 & 55.67 $\pm$ 1.19 & 0.042 $\pm$ 0.016 & 49.81 $\pm$ 0.41 & 50.12 $\pm$ 0.15 \\
BV-fam-S00550001 & 31 & 10,952 & 57.29 $\pm$ 2.22 & 0.050 $\pm$ 0.017 & 50.04 $\pm$ 0.28 & 50.10 $\pm$ 0.41 \\
BV-fam-S00340002 & 7 & 3,457 & 61.83 $\pm$ 1.23 & 0.052 $\pm$ 0.015 & 49.85 $\pm$ 0.20 & 50.06 $\pm$ 0.26 \\
BV-fam-S00350001 & 25 & 2,959 & 56.70 $\pm$ 2.17 & 0.006 $\pm$ 0.009 & 49.86 $\pm$ 0.33 & 49.97 $\pm$ 0.17 \\
BV-fam-S01010001 & 23 & 1,170 & 55.19 $\pm$ 2.94 & 0.014 $\pm$ 0.018 & 49.97 $\pm$ 0.33 & 50.21 $\pm$ 0.34 \\
\bottomrule
\end{tabular}}
\caption{Individual-family BabyView results for GPT-BERT base with seed-level variability, sorted by training tokens. Age is months at onboarding time of that family's given child.}\label{tab:ind_family_results_gptbert_base}
\end{table}

\begin{table}[t]
\centering
\resizebox{\textwidth}{!}{\begin{tabular}{lrrrrrr}
\toprule
Family & Age (months) & Tokens & Zorro & WordSim & COMPS & EWoK \\
\midrule
BV-fam-S00400001 & 31 & 725,600 & 54.37 $\pm$ 1.63 & 0.072 $\pm$ 0.016 & 50.10 $\pm$ 0.10 & 50.00 $\pm$ 0.11 \\
BV-fam-S00510002 & 11 & 569,788 & 55.65 $\pm$ 0.85 & 0.068 $\pm$ 0.047 & 50.12 $\pm$ 0.33 & 49.96 $\pm$ 0.20 \\
BV-fam-S00360001 & 8 & 257,817 & 57.97 $\pm$ 1.34 & 0.066 $\pm$ 0.021 & 50.22 $\pm$ 0.23 & 49.88 $\pm$ 0.39 \\
BV-fam-S00220001 & 5 & 242,954 & 57.24 $\pm$ 1.12 & 0.074 $\pm$ 0.018 & 49.98 $\pm$ 0.36 & 50.17 $\pm$ 0.34 \\
BV-fam-S00400002 & 28 & 181,317 & 59.01 $\pm$ 1.77 & 0.058 $\pm$ 0.016 & 49.77 $\pm$ 0.23 & 50.06 $\pm$ 0.28 \\
BV-fam-S00320002 & 7 & 171,656 & 56.21 $\pm$ 1.77 & 0.056 $\pm$ 0.019 & 50.07 $\pm$ 0.27 & 49.99 $\pm$ 0.26 \\
BV-fam-S00370001 & 8 & 167,136 & 57.87 $\pm$ 0.77 & 0.066 $\pm$ 0.023 & 50.13 $\pm$ 0.25 & 50.01 $\pm$ 0.10 \\
BV-fam-S00430001 & 31 & 144,253 & 56.96 $\pm$ 1.25 & 0.064 $\pm$ 0.018 & 50.08 $\pm$ 0.24 & 50.05 $\pm$ 0.14 \\
BV-fam-S00510001 & 11 & 131,120 & 58.44 $\pm$ 2.12 & 0.072 $\pm$ 0.023 & 50.03 $\pm$ 0.16 & 50.05 $\pm$ 0.21 \\
BV-fam-S00720001 & 38 & 100,807 & 57.60 $\pm$ 1.36 & 0.050 $\pm$ 0.029 & 49.95 $\pm$ 0.21 & 50.13 $\pm$ 0.29 \\
BV-fam-S00490001 & 27 & 40,835 & 56.85 $\pm$ 0.92 & 0.032 $\pm$ 0.029 & 50.01 $\pm$ 0.26 & 50.06 $\pm$ 0.23 \\
BV-fam-S00230001 & 22 & 26,501 & 57.69 $\pm$ 2.79 & 0.038 $\pm$ 0.025 & 50.05 $\pm$ 0.24 & 49.99 $\pm$ 0.13 \\
BV-fam-S00440001 & 26 & 21,313 & 55.95 $\pm$ 2.04 & -0.004 $\pm$ 0.015 & 49.77 $\pm$ 0.26 & 49.94 $\pm$ 0.20 \\
BV-fam-S00360002 & 8 & 15,006 & 54.91 $\pm$ 1.44 & 0.036 $\pm$ 0.017 & 50.14 $\pm$ 0.20 & 49.93 $\pm$ 0.26 \\
BV-fam-S00460001 & 11 & 14,618 & 56.60 $\pm$ 2.05 & 0.062 $\pm$ 0.023 & 50.11 $\pm$ 0.19 & 49.92 $\pm$ 0.15 \\
BV-fam-S00350002 & 8 & 12,614 & 56.14 $\pm$ 1.43 & 0.026 $\pm$ 0.030 & 50.13 $\pm$ 0.34 & 49.80 $\pm$ 0.17 \\
BV-fam-S00550001 & 31 & 10,952 & 58.51 $\pm$ 2.10 & 0.034 $\pm$ 0.013 & 49.88 $\pm$ 0.28 & 49.87 $\pm$ 0.41 \\
BV-fam-S00340002 & 7 & 3,457 & 59.41 $\pm$ 1.92 & 0.052 $\pm$ 0.022 & 49.87 $\pm$ 0.19 & 50.01 $\pm$ 0.19 \\
BV-fam-S00350001 & 25 & 2,959 & 56.17 $\pm$ 3.85 & 0.028 $\pm$ 0.011 & 49.99 $\pm$ 0.33 & 50.04 $\pm$ 0.17 \\
BV-fam-S01010001 & 23 & 1,170 & 55.31 $\pm$ 0.93 & 0.032 $\pm$ 0.028 & 50.18 $\pm$ 0.25 & 49.78 $\pm$ 0.06 \\
\bottomrule
\end{tabular}}
\caption{Individual-family BabyView results for GPT-BERT small with seed-level variability, sorted by training tokens. Age is months at onboarding time of that family's given child.}\label{tab:ind_family_results_gptbert_small}
\end{table}

\section{Individual BabyView Family Experiment Results}\label{app:ind_family_results}\label{app:ind_family_results}

See Tables \ref{tab:ind_family_results_gpt2_small} to \ref{tab:ind_family_results_gptbert_small} for BabyView evaluation results of the individual family models.

\begin{small}
\setlength{\tabcolsep}{4pt}
\renewcommand{\arraystretch}{1.05}

\begin{table}[t]
\centering

\begin{tabular}{@{}llcrrrrr@{}}
\toprule
\textbf{Model} & \textbf{Target} & \textbf{$\alpha$} & \makecell{\textbf{NZ}\\\textbf{feat.}} & \makecell{\textbf{Lasso}\\\textbf{$R^2$}} & \makecell{\textbf{Lasso}\\\textbf{RMSE}} & \makecell{\textbf{XGB}\\\textbf{$R^2$}} & \makecell{\textbf{XGB}\\\textbf{RMSE}} \\
\midrule

GPT-2 Mini & Zorro & 0.0488 & 21 & 0.657 & 1.784 & 0.395 & 2.369 \\
GPT-2 Mini & WordSim & 0.0026 & 13 & -0.208 & 0.029 & -0.166 & 0.029 \\
GPT-2 Mini & COMPS & 0.0668 & 0 & -0.144 & 0.166 & -1.076 & 0.223 \\

\midrule

GPT-2 Small & Zorro & 0.5230 & 6 & 0.289 & 2.974 & 0.172 & 3.208 \\
GPT-2 Small & WordSim & 0.0173 & 1 & -1.047 & 0.031 & -1.943 & 0.037 \\
GPT-2 Small & COMPS & 0.0713 & 3 & -1.479 & 0.257 & -0.601 & 0.207 \\

\midrule

GPT-BERT Base & Zorro & 0.3479 & 8 & -0.164 & 2.675 & -0.415 & 2.949 \\
GPT-BERT Base & WordSim & 0.0022 & 10 & 0.456 & 0.026 & 0.137 & 0.032 \\
GPT-BERT Base & COMPS & 0.0191 & 13 & -0.117 & 0.236 & -0.683 & 0.290 \\

\midrule

GPT-BERT Small & Zorro & 1.0621 & 0 & -0.954 & 2.558 & -0.386 & 2.154 \\
GPT-BERT Small & WordSim & 0.0028 & 10 & 0.109 & 0.019 & 0.182 & 0.018 \\
GPT-BERT Small & COMPS & 0.0707 & 1 & -0.159 & 0.171 & -0.592 & 0.200 \\

\bottomrule
\end{tabular}

\caption{
Cross-validated predictive performance of lasso regression and XGBoost models for each model–target pair. 
$\alpha$ denotes the lasso regularization strength selected via cross-validation. 
NZ feat. is the number of features with nonzero coefficients in the fitted lasso model, reflecting model sparsity. 
Lasso $R^2$ and XGB $R^2$ report the proportion of variance explained on held-out data (higher is better), 
while Lasso RMSE and XGB RMSE report root mean squared error (lower is better). 
Together, these metrics characterize both predictive performance and model complexity across methods.
}

\label{app_tab:predictive_model_metrics}

\end{table}
\end{small}

\section{LLM Usage Disclosure}

We used LLMs (e.g., GPT-5) to assist with parts of the coding process and limited aspects of paper preparation, including LaTeX table formatting and minor editing for clarity and grammar. All outputs were carefully reviewed to ensure accuracy and appropriateness.


\section{Linguistic Analysis: All Linguistic Features \& Results}\label{app:full_linguistic_features}

See Table \ref{app_tab:predictive_model_metrics} for cross-validated Lasso and XGBoost predictive performance of our models on each target benchmark (Zorro, WordSim, COMPS). We see that our linguistic features can explain Zorro performance reasonably well for some models, but results are inconsistent across architectures. WordSim is only partially explained (primarily for GPT-BERT models), while COMPS shows low between-experiment variance and remains difficult to explain from the current feature set. See Table \ref{app_tab:ling_feature_catalog_detailed} for a full catalog (list and description) of all 175 linguistic features we considered for our analysis, and Table \ref{app_tab:ling_full_grouped} for statistical analysis results of all 175 linguistic features.

\begin{scriptsize}
\setlength{\tabcolsep}{4pt}
\setlength{\LTleft}{0pt}
\setlength{\LTright}{0pt}
\renewcommand{\arraystretch}{1.02}
\newcommand{\featureline}{\specialrule{0.15pt}{0pt}{0pt}}

\begin{longtable}{>{\raggedright\arraybackslash}p{0.34\textwidth}>{\raggedright\arraybackslash}p{0.61\textwidth}}
\caption{Comprehensive catalog of the 175 linguistic features considered in our BabyView analysis. The table is grouped by feature family and provides a one-sentence description of each feature. Note that all 175 features were kept after restricting to numeric variables and applying missingness ($\leq 0.4$ across the 25 experiments) and zero-variance filtering. Hence, all 175 features were used in the final statistical analyses (Spearman, lasso, and XGBoost).}\label{app_tab:ling_feature_catalog_detailed}\\
\toprule
\textbf{Feature name} & \textbf{Brief description} \\
\midrule
\endfirsthead
\toprule
\textbf{Feature name} & \textbf{Brief description} \\
\midrule
\endhead
\multicolumn{2}{@{}l}{\textbf{Lexical / Distributional / Scale}} \\
\midrule
avg\_conversation\_len\_tokens & This feature measures the average number of tokens per conversation, capturing how much language is typically contained in a single interaction. \\
\featureline
avg\_sentence\_len\_tokens & This feature measures the average number of tokens per sentence, providing a simple proxy for sentence-level length and complexity. \\
\featureline
avg\_utterance\_len\_tokens & This feature measures the average number of tokens per utterance, summarizing how long individual turns tend to be. \\
\featureline
bigram\_entropy & This feature measures the entropy of the bigram distribution, so higher values indicate greater diversity in adjacent token sequences. \\
\featureline
bigram\_mutual\_information & This feature measures the average pointwise mutual information of adjacent token pairs, so higher values indicate that neighboring words are more strongly associated than expected by chance. \\
\featureline
burstiness\_measure & This feature measures the variability of utterance lengths relative to their mean, capturing how bursty or uneven the discourse is. \\
\featureline
character\_entropy & This feature measures the entropy of the non-whitespace character distribution, capturing low-level orthographic diversity. \\
\featureline
compression\_ratio & This feature measures the ratio of compressed to uncompressed text length, serving as a rough proxy for redundancy and repetitiveness in the corpus. \\
\featureline
\makecell[l]{conversations\_lt\_128\_tokens\\pct} & This feature measures the proportion of conversations shorter than 128 tokens, capturing the prevalence of very short interactions. \\
\featureline
\makecell[l]{conversations\_lt\_512\_tokens\\pct} & This feature measures the proportion of conversations shorter than 512 tokens, capturing how concentrated the dataset is in relatively short interactions. \\
\featureline
diminutive\_frequency & This feature summarizes the corpus property captured by diminutive frequency. \\
\featureline
dis\_legomena\_ratio & This feature measures the proportion of vocabulary types that occur exactly twice in the corpus, providing a second view of lexical tail structure beyond hapax items. \\
\featureline
elongated\_word\_frequency & This feature measures how often words with elongated character sequences appear, capturing expressive spellings such as stretched vowels or consonants. \\
\featureline
emotion\_word\_pct & This feature measures the proportion of utterances containing words from an emotion-related lexicon, capturing how frequently affective vocabulary appears in the corpus. \\
\featureline
eventive\_verb\_pct & This feature summarizes the corpus property captured by eventive verb pct. \\
\featureline
eventive\_verb\_token\_pct & This feature measures the proportion of tokens that belong to an eventive-verb lexicon, capturing how much the corpus emphasizes actions and events. \\
\featureline
\makecell[l]{frequency\_distribution\\skewness} & This feature measures the skewness of the token-frequency distribution, capturing how strongly the corpus is dominated by a small number of very frequent words. \\
\featureline
hapax\_ratio & This feature measures the proportion of vocabulary types that occur exactly once in the corpus, capturing how much lexical mass lies in the long tail. \\
\featureline
interjection\_frequency & This feature measures how often interjections such as 'oh' or 'wow' appear, capturing expressive and affective discourse style. \\
\featureline
intrinsic\_perplexity\_proxy & This feature is a simple intrinsic predictability proxy derived from token entropy, with higher values indicating a less predictable token distribution. \\
\featureline
lexical\_diversity\_slope & This feature measures how lexical diversity changes over time across the dataset, capturing whether later conversations become more or less varied in word choice. \\
\featureline
mattr & This feature measures moving-average type-token ratio over a sliding window, giving a more length-robust estimate of lexical diversity than raw TTR. \\
\featureline
mean\_utterance\_length\_slope & This feature measures the linear change in average utterance length over the course of the dataset, capturing whether turns become longer or shorter over time. \\
\featureline
\makecell[l]{narrative\_discourse\_marker\\pct} & This feature measures the proportion of utterances containing narrative discourse markers such as temporal or sequencing words, capturing how often the corpus uses narrative-style structuring. \\
\featureline
\makecell[l]{narrative\_discourse\_marker\\token\_pct} & This feature measures the proportion of tokens that are narrative discourse markers, capturing the token-level prevalence of narrative-style sequencing language. \\
\featureline
num\_conversations & This feature counts the number of conversations in the dataset, providing a coarse measure of dataset coverage and interactional breadth. \\
\featureline
num\_sentences & This feature summarizes the corpus property captured by num sentences. \\
\featureline
num\_tokens & This feature counts the total number of tokens in the corpus, providing a direct measure of dataset size. \\
\featureline
num\_types & This feature counts the number of unique token types in the corpus, providing a basic measure of vocabulary breadth. \\
\featureline
num\_utterances & This feature counts the number of utterances in the corpus, providing a basic measure of conversational segmentation and turn volume. \\
\featureline
past\_tense\_verb\_pct & This feature measures the average per-utterance rate of past-tense verbs, capturing how often speakers refer to past events. \\
\featureline
past\_tense\_verb\_token\_pct & This feature measures the proportion of tokens that are identified as past-tense verbs by a simple heuristic, capturing past-event language at the token level. \\
\featureline
positive\_sentiment\_word\_pct & This feature summarizes the corpus property captured by positive sentiment word pct. \\
\featureline
praise\_word\_pct & This feature summarizes the corpus property captured by praise word pct. \\
\featureline
sound\_effect\_frequency & This feature measures how often onomatopoeic or sound-effect words occur, capturing playful and imitative sound language. \\
\featureline
\makecell[l]{syntactic\_complexity\_slope\\over\_time\_proxy} & This feature measures the change in a proxy for syntactic complexity over time, operationalized from shifts in average utterance length across the dataset. \\
\featureline
token\_entropy & This feature measures the entropy of the unigram token distribution, with higher values indicating a more diverse and less concentrated vocabulary. \\
\featureline
top1000\_token\_coverage & This feature measures the proportion of all tokens covered by the 1,000 most frequent word types, capturing how concentrated the corpus is in a relatively small core vocabulary. \\
\featureline
top100\_token\_proportion & This feature measures the proportion of all tokens covered by the 100 most frequent word types, capturing lexical concentration among the most common words. \\
\featureline
trigram\_entropy & This feature measures the entropy of the trigram distribution, so higher values indicate greater diversity and less predictability in local three-token sequences. \\
\featureline
ttr & This feature measures the raw type-token ratio, or the number of unique tokens divided by total tokens, as a simple index of lexical diversity. \\
\featureline
unigram\_entropy & This feature measures the entropy of the unigram token distribution, with higher values indicating a more diverse and less predictable vocabulary. \\
\featureline
\makecell[l]{vocabulary\_growth\_over\_time\\slope} & This feature measures the rate at which new vocabulary accumulates over the course of the dataset, capturing how quickly lexical novelty is introduced. \\
\featureline
zipf\_slope & This feature measures the slope of the log-rank versus log-frequency relationship, capturing how closely the vocabulary frequency distribution follows a Zipf-like pattern. \\
\featureline
\midrule
\multicolumn{2}{@{}l}{\textbf{POS / Syntactic Composition}} \\
\midrule
\makecell[l]{dep\_avg\_clause\_count\_per\\sentence} & This feature measures the average number of clause-related dependency relations per sentence, capturing how clause-dense the parsed language is. \\
\featureline
\makecell[l]{dep\_avg\_clause\_dependency\\count\_per\_sentence\_proxy} & This feature is a proxy for clause density based on clause-related dependency relations per sentence. \\
\featureline
dep\_avg\_dependency\_length & This feature measures the average linear distance between tokens and their syntactic heads, a standard proxy for syntactic locality or complexity. \\
\featureline
dep\_avg\_parse\_tree\_depth & This feature measures average dependency parse-tree depth, which serves as a structural proxy for syntactic complexity. \\
\featureline
\makecell[l]{dep\_avg\_parse\_tree\_depth\\proxy} & This feature is a proxy version of average parse-tree depth used for analyses that rely on a compact syntactic-complexity summary. \\
\featureline
\makecell[l]{dep\_dependency\_length\\variance} & This feature measures the variance of dependency lengths, capturing how variable syntactic attachment distances are across the corpus. \\
\featureline
dep\_left\_branching\_ratio & This feature measures the proportion of dependency links that branch to the left, capturing one aspect of syntactic branching direction. \\
\featureline
dep\_num\_docs\_parsed & This feature counts how many utterance documents were successfully parsed, providing a measure of usable input for the dependency parser. \\
\featureline
dep\_num\_sentences & This feature counts the number of parsed sentences, providing the sample size for sentence-level dependency metrics. \\
\featureline
dep\_pct\_coordinate\_clauses & This feature measures the proportion of parsed sentences containing coordination, capturing how often clauses are linked side by side. \\
\featureline
dep\_pct\_fragments\_parse & This feature measures the proportion of parsed sentences classified as fragments, capturing incomplete or very short syntactic units. \\
\featureline
dep\_pct\_imperatives\_parse & This feature measures the proportion of parsed sentences classified as imperatives, capturing directive clause structure. \\
\featureline
\makecell[l]{dep\_pct\_passive\\constructions} & This feature measures the proportion of parsed sentences containing passive markers, capturing how often passive syntax appears. \\
\featureline
dep\_pct\_questions\_parse & This feature measures the proportion of parsed sentences identified as questions, using the dependency-parse pipeline rather than simple punctuation alone. \\
\featureline
dep\_pct\_subordinate\_clauses & This feature measures the proportion of parsed sentences containing subordinate clause relations, capturing hierarchical syntactic embedding. \\
\featureline
dep\_pct\_wh\_questions\_parse & This feature measures the proportion of parsed sentences that are WH-questions, capturing interrogatives introduced by words such as 'what' or 'where'. \\
\featureline
\makecell[l]{dep\_pct\_yes\_no\_questions\\parse} & This feature measures the proportion of parsed sentences that are yes/no questions, capturing interrogatives formed without WH-words. \\
\featureline
dep\_right\_branching\_ratio & This feature measures the proportion of dependency links that branch to the right, capturing one aspect of syntactic branching direction. \\
\featureline
\makecell[l]{dep\_total\_utterances\\available} & This feature counts the total number of utterances that were available before dependency parsing, providing a measure of syntactic evidence volume. \\
\featureline
\makecell[l]{dep\_utterances\_used\_for\\parsing} & This feature counts how many utterances were actually passed to the dependency parser after any sampling or filtering. \\
\featureline
pos\_num\_tagged\_tokens & This feature counts the total number of POS-tagged tokens across the corpus, providing the amount of text included in the POS analysis pipeline. \\
\featureline
\makecell[l]{pos\_num\_tagged\_tokens\\caregiver} & This feature counts the number of POS-tagged tokens produced by the caregiver, capturing how much caregiver speech is available for part-of-speech analysis. \\
\featureline
pos\_num\_tagged\_tokens\_child & This feature counts the number of POS-tagged tokens produced by the child, capturing how much child speech is available for part-of-speech analysis. \\
\featureline
pos\_pct\_adjectives & This feature measures the proportion of POS-tagged tokens that are adjectives, capturing how much descriptive modifier language appears. \\
\featureline
pos\_pct\_adverbs & This feature measures the proportion of POS-tagged tokens that are adverbs, capturing manner, degree, and discourse-modifying language. \\
\featureline
pos\_pct\_conjunctions & This feature measures the proportion of POS-tagged tokens that are conjunctions, capturing how often clauses or phrases are linked together. \\
\featureline
pos\_pct\_content\_words & This feature measures the proportion of POS-tagged tokens that are content words, capturing how much of the corpus consists of semantically rich lexical items. \\
\featureline
pos\_pct\_determiners & This feature measures the proportion of POS-tagged tokens that are determiners, capturing the use of noun-phrase marking words such as articles and demonstratives. \\
\featureline
pos\_pct\_function\_words & This feature measures the proportion of POS-tagged tokens that are function words, capturing closed-class grammatical material. \\
\featureline
pos\_pct\_nouns & This feature measures the proportion of POS-tagged tokens that are nouns, capturing how strongly the corpus emphasizes object and entity reference. \\
\featureline
pos\_pct\_nouns\_caregiver & This feature measures the proportion of caregiver POS-tagged tokens that are nouns, capturing noun usage specifically in caregiver speech. \\
\featureline
pos\_pct\_nouns\_child & This feature measures the proportion of child POS-tagged tokens that are nouns, capturing noun usage specifically in child speech. \\
\featureline
pos\_pct\_prepositions & This feature measures the proportion of POS-tagged tokens that are prepositions or adpositions, capturing relational language. \\
\featureline
pos\_pct\_pronouns & This feature measures the proportion of POS-tagged tokens that are pronouns, capturing referential language that depends on discourse context. \\
\featureline
pos\_pct\_verbs & This feature measures the proportion of POS-tagged tokens that are verbs, capturing the amount of event- and action-centered language. \\
\featureline
pos\_pct\_verbs\_caregiver & This feature measures the proportion of caregiver POS-tagged tokens that are verbs, capturing verb usage specifically in caregiver speech. \\
\featureline
pos\_pct\_verbs\_child & This feature measures the proportion of child POS-tagged tokens that are verbs, capturing verb usage specifically in child speech. \\
\featureline
pos\_pos\_bigram\_diversity & This feature measures the diversity of observed POS bigrams relative to the number of POS transitions, capturing local syntactic variety. \\
\featureline
\makecell[l]{pos\_pos\_bigram\_diversity\\tagspace} & This feature measures POS-bigram diversity normalized by the size of the possible tag space, capturing how much of the available local syntactic space is used. \\
\featureline
pos\_pos\_bigram\_entropy & This feature measures the entropy of adjacent POS-tag bigrams, capturing the diversity of local syntactic patterns in the corpus. \\
\featureline
pos\_pos\_entropy & This feature measures entropy over POS tags, capturing how evenly different part-of-speech categories are distributed. \\
\featureline
pos\_pos\_entropy\_caregiver & This feature measures POS-tag entropy in caregiver speech, capturing syntactic-category diversity specifically for caregiver language. \\
\featureline
pos\_pos\_entropy\_child & This feature measures POS-tag entropy in child speech, capturing syntactic-category diversity specifically for child language. \\
\featureline
\makecell[l]{pos\_total\_utterances\\available} & This feature counts the total number of utterances available before the POS-tagging pipeline, providing the available sample size for POS analysis. \\
\featureline
pos\_utterances\_used\_for\_pos & This feature counts how many utterances were actually used for POS tagging after any sampling or filtering. \\
\featureline
\midrule
\multicolumn{2}{@{}l}{\textbf{Conversational / Discourse Structure}} \\
\midrule
\makecell[l]{adjacent\_turn\_lexical\\overlap} & This feature measures lexical overlap between adjacent turns, capturing how much consecutive speakers reuse the same words. \\
\featureline
anaphora\_usage\_pct & This feature measures the proportion of tokens that are third-person anaphoric forms such as pronouns, capturing discourse-dependent reference. \\
\featureline
avg\_caregiver\_utterance\_len & This feature measures the average length of caregiver utterances in tokens, summarizing how long caregiver turns tend to be. \\
\featureline
avg\_child\_utterance\_len & This feature measures the average length of child utterances in tokens, summarizing how long child turns tend to be. \\
\featureline
avg\_conversation\_len\_turns & This feature measures the average number of turns per conversation, capturing interactional depth independent of token length. \\
\featureline
\makecell[l]{caregiver\_after\_child\_any\\overlap\_pct} & This feature measures the proportion of child-to-caregiver adjacent pairs in which the caregiver reuses at least some child vocabulary. \\
\featureline
caregiver\_child\_token\_ratio & This feature measures the ratio of caregiver tokens to child tokens, summarizing the relative contribution of the two speaker groups. \\
\featureline
caregiver\_question\_pct & This feature measures the proportion of caregiver utterances that are questions, capturing how often caregivers prompt or query the child. \\
\featureline
\makecell[l]{caregiver\_reuse\_child\_tokens\\pct} & This feature measures the proportion of child-to-caregiver adjacent pairs in which the caregiver reuses child vocabulary, operationalized as lexical overlap greater than zero. \\
\featureline
caregiver\_token\_pct & This feature measures the proportion of all corpus tokens produced by caregivers. \\
\featureline
caregiver\_wh\_question\_pct & This feature measures the proportion of caregiver utterances that are WH-questions, capturing a more specific type of information-seeking prompt. \\
\featureline
child\_token\_pct & This feature measures the proportion of all corpus tokens produced by the child. \\
\featureline
clarification\_pct & This feature measures the proportion of utterances containing clarification markers or repair-oriented questions, capturing requests for clearer information. \\
\featureline
confirmation\_pct & This feature measures the proportion of utterances containing confirmation markers such as 'yes' or 'okay', capturing explicit acknowledgement or agreement. \\
\featureline
directive\_pct & This feature measures the proportion of utterances classified as directives or commands, capturing instruction-like language. \\
\featureline
expansion\_pct & This feature measures the proportion of utterances that contain expansion markers suggesting elaboration or reformulation of prior content. \\
\featureline
fragment\_pct & This feature measures the proportion of utterances classified as fragments, typically because they are very short or incomplete. \\
\featureline
imperative\_pct & This feature measures the proportion of utterances classified as imperatives, capturing command-like speech. \\
\featureline
\makecell[l]{overlapping\_speech\_marker\\pct} & This feature measures the rate of overlap markers in the transcript, capturing instances where speech overlaps or is explicitly annotated as overlapping. \\
\featureline
pronoun\_density & This feature measures the proportion of all tokens that are pronouns, capturing the density of discourse-dependent reference. \\
\featureline
question\_pct & This feature measures the proportion of utterances that are questions, based primarily on question-mark cues. \\
\featureline
self\_repair\_frequency & This feature measures how often utterances contain fillers or repair markers such as restarts, corrections, or hesitation expressions. \\
\featureline
speaker\_label\_consistency & This feature measures the proportion of utterances whose speaker labels are recognized and consistent with the expected label set. \\
\featureline
\makecell[l]{topic\_entropy\_overlap\\heuristic\_proxy} & This feature is an overlap-based proxy for topical variation, using a coarse heuristic intended to capture topic dispersion rather than a full topic model. \\
\featureline
\makecell[l]{topic\_persistence\_length\\overlap\_heuristic\_proxy} & This feature measures the average run length of local lexical-overlap segments, capturing how long a topic tends to persist across turns. \\
\featureline
turn\_taking\_frequency & This feature measures how often speaker identity switches across adjacent turns, capturing the pace of turn alternation. \\
\featureline
\makecell[l]{unknown\_speaker\_utterance\\pct} & This feature measures the proportion of utterances with unknown or unrecognized speaker labels. \\
\featureline
wh\_question\_pct & This feature measures the proportion of utterances that are WH-questions, capturing interrogatives introduced by question words. \\
\featureline
yes\_no\_question\_pct & This feature measures the proportion of utterances that are yes/no questions rather than WH-questions. \\
\featureline
\midrule
\multicolumn{2}{@{}l}{\textbf{Semantic Coherence}} \\
\midrule
\makecell[l]{sem\_mean\_adjacent\_turn\\semantic\_similarity} & This feature measures the average embedding-based semantic similarity between adjacent turns, capturing how semantically connected neighboring utterances are. \\
\featureline
\makecell[l]{sem\_mean\_adjacent\_turn\\semantic\_similarity\_len\\weighted} & This feature measures the average semantic similarity between adjacent turns while weighting pairs by turn length, so longer turns contribute more to the summary. \\
\featureline
\makecell[l]{sem\_mean\_semantic\_similarity\\caregiver\_to\_child} & This feature measures the average semantic similarity from caregiver turns to immediately following child turns, capturing one direction of interactional semantic alignment. \\
\featureline
\makecell[l]{sem\_mean\_semantic\_similarity\\child\_to\_caregiver} & This feature measures the average semantic similarity from child turns to immediately following caregiver turns, capturing the reverse direction of interactional semantic alignment. \\
\featureline
sem\_num\_adjacent\_pairs\_used & This feature counts the number of adjacent utterance pairs included in the semantic-similarity computation after filtering and sampling. \\
\featureline
\makecell[l]{sem\_num\_pairs\_caregiver\_to\\child} & This feature counts the number of caregiver-to-child adjacent turn pairs available for directional semantic-similarity analysis. \\
\featureline
\makecell[l]{sem\_num\_pairs\_child\_to\\caregiver} & This feature counts the number of child-to-caregiver adjacent turn pairs available for directional semantic-similarity analysis. \\
\featureline
\makecell[l]{sem\_total\_adjacent\_pairs\\available} & This feature counts all adjacent utterance pairs available before semantic-similarity sampling or filtering. \\
\featureline
\makecell[l]{sem\_total\_utterances\\available} & This feature counts the total number of utterances available before the semantic-similarity pipeline is applied. \\
\featureline
\makecell[l]{sem\_utterances\_used\_for\\semantic} & This feature counts how many utterances were actually used when constructing adjacent pairs for semantic analysis. \\
\featureline
\midrule
\multicolumn{2}{@{}l}{\textbf{Repetition / Formulaicity}} \\
\midrule
\makecell[l]{exact\_sentence\_repetition\\rate} & This feature measures how often whole utterances repeat exactly, capturing formulaic reuse at the sentence level. \\
\featureline
ngram\_repetition\_rate & This feature measures the overall rate of repeated bigrams and trigrams, capturing formulaicity and local repetition in the corpus. \\
\featureline
partial\_repetition\_frequency & This feature measures how often utterances contain partial repetition or reduplication-like patterns, capturing local repetition within turns. \\
\featureline
reduplication\_frequency & This feature measures how often explicit reduplication patterns occur, such as repeated words or hyphenated duplicates. \\
\featureline
repeated\_bigram\_ratio & This feature measures the proportion of all bigram tokens that belong to repeated bigram types, capturing repeated two-word sequences. \\
\featureline
repeated\_trigram\_ratio & This feature measures the proportion of all trigram tokens that belong to repeated trigram types, capturing repeated three-word sequences. \\
\featureline
\midrule
\multicolumn{2}{@{}l}{\textbf{Mixture / Composition Metadata}} \\
\midrule
\makecell[l]{age\_age\_metadata\_alignment\\score} & This feature summarizes how tightly concentrated the ages of the component families are within a mixture, with higher values indicating stronger age alignment. \\
\featureline
age\_num\_component\_families & This feature counts how many component families contribute to a mixture in the age-metadata analysis. \\
\featureline
age\_num\_families\_with\_age & This feature counts how many component families in a mixture have usable age metadata available. \\
\featureline
age\_weighted\_age\_std\_months & This feature measures the weighted standard deviation of component-family ages in months, capturing age heterogeneity within a mixture. \\
\featureline
age\_weighted\_mean\_age\_months & This feature measures the weighted mean age in months of component families contributing to a mixture. \\
\featureline
div\_max\_js\_with\_others & This feature measures the maximum Jensen–Shannon divergence between a dataset and any other dataset, capturing its strongest distributional distinctiveness. \\
\featureline
div\_mean\_js\_with\_others & This feature measures the mean Jensen–Shannon divergence between a dataset and all other datasets, capturing average distributional distinctiveness. \\
\featureline
div\_mean\_kl\_from\_others & This feature measures the mean KL divergence from other datasets into the current one, capturing how distinct the dataset is when others are compared against it. \\
\featureline
div\_mean\_kl\_to\_others & This feature measures the mean KL divergence from the current dataset to other datasets, capturing how distinct its token distribution is relative to the rest. \\
\featureline
div\_median\_js\_with\_others & This feature measures the median Jensen–Shannon divergence between a dataset and all others, providing a robust summary of distributional distinctiveness. \\
\featureline
\makecell[l]{mix\_content\_vocab\_support\\size} & This feature counts the size of the support of content-word vocabulary in the mixture distribution after formatting and non-content tokens are removed. \\
\featureline
\makecell[l]{mix\_cross\_family\_lexical\\overlap} & This feature measures average lexical overlap across component families in a mixture, capturing how similar their vocabularies are. \\
\featureline
\makecell[l]{mix\_effective\_vocab\_size\\bits} & This feature converts mixture token entropy in bits into an effective vocabulary size, summarizing how broadly probability mass is spread across word types. \\
\featureline
\makecell[l]{mix\_effective\_vocab\_size\\content\_only} & This feature is the effective vocabulary size computed after removing formatting-style tokens, capturing lexical breadth in content words only. \\
\featureline
\makecell[l]{mix\_effective\_vocab\_size\\nats} & This feature converts mixture token entropy in nats into an effective vocabulary size, providing an alternative entropy-scale summary of lexical breadth. \\
\featureline
\makecell[l]{mix\_effective\_vocabulary\\size} & This feature summarizes the effective number of vocabulary items represented in the mixture distribution based on entropy. \\
\featureline
mix\_family\_diversity\_index & This feature measures how evenly a mixture is distributed across its component families, with higher values indicating a more balanced mixture. \\
\featureline
mix\_formatting\_token\_mass & This feature measures how much probability mass in the mixture distribution is assigned to formatting or metadata-like tokens. \\
\featureline
\makecell[l]{mix\_inter\_family\_variance\\entropy} & This feature measures variance in per-family token entropy within a mixture, capturing heterogeneity in lexical concentration across families. \\
\featureline
\makecell[l]{mix\_inter\_family\_variance\\syntactic\_complexity} & This feature measures variance in a syntactic-complexity proxy across families within a mixture, capturing cross-family heterogeneity in utterance structure. \\
\featureline
\makecell[l]{mix\_js\_divergence\_within\\mixture} & This feature measures average pairwise Jensen–Shannon divergence among component families within a mixture, capturing internal distributional heterogeneity. \\
\featureline
\makecell[l]{mix\_mixture\_token\_entropy\\bits} & This feature measures the entropy of the mixture token distribution in bits, capturing how diverse and even the overall mixture vocabulary is. \\
\featureline
\makecell[l]{mix\_mixture\_token\_entropy\\nats} & This feature measures the entropy of the mixture token distribution in natural-log units, providing the same information on a different scale. \\
\featureline
\makecell[l]{mix\_mixture\_top100\_token\\mass} & This feature measures the fraction of mixture probability mass captured by the 100 most probable tokens, reflecting lexical concentration. \\
\featureline
mix\_mixture\_top10\_token\_mass & This feature measures the fraction of mixture probability mass captured by the 10 most probable tokens, reflecting lexical concentration in a small core vocabulary. \\
\featureline
mix\_mixture\_top1\_token\_mass & This feature measures the fraction of mixture probability mass assigned to the single most probable token. \\
\featureline
mix\_mixture\_top50\_token\_mass & This feature measures the fraction of mixture probability mass captured by the 50 most probable tokens. \\
\featureline
mix\_mixture\_top5\_token\_mass & This feature measures the fraction of mixture probability mass captured by the 5 most probable tokens. \\
\featureline
\makecell[l]{mix\_mixture\_vocab\_support\\size} & This feature counts how many token types have nonzero probability in the overall mixture distribution. \\
\featureline
\makecell[l]{mix\_mixture\_weight\_entropy\\bits} & This feature measures the entropy of the component-family mixture weights, capturing how evenly the mixture draws from its sources. \\
\featureline
mix\_num\_component\_families & This feature counts the number of component families contributing nonzero weight to a mixture. \\
\featureline
\makecell[l]{mix\_token\_entropy\_bits\\content\_only} & This feature measures token entropy in bits after removing formatting-like tokens, capturing content-word diversity. \\
\featureline
mix\_vocab\_support\_size & This feature counts how many token types have nonzero probability in the mixture and functions as a support-size summary of vocabulary breadth. \\
\featureline
\midrule
\multicolumn{2}{@{}l}{\textbf{Reference LM Predictability}} \\
\midrule
\makecell[l]{ppl\_norm\_babylm20m\_caregiver\\only} & This feature measures normalized perplexity under a BabyLM-trained reference language model on caregiver-only text, capturing how predictable the caregiver stream is to that model. \\
\featureline
\makecell[l]{ppl\_norm\_babylm20m\_full\\dialogue} & This feature measures normalized perplexity under a BabyLM-trained reference language model on the full dialogue stream, capturing overall corpus predictability. \\
\featureline
\makecell[l]{ppl\_norm\_babylm20m\_mean\\scopes} & This feature averages the BabyLM reference-model perplexity summaries across scoring scopes, providing a broader predictability estimate. \\
\featureline
\makecell[l]{ppl\_norm\_external\_gpt2\\caregiver\_only} & This feature measures normalized perplexity under an external GPT-2 model on caregiver-only text, capturing how predictable caregiver language is to a general-domain model. \\
\featureline
\makecell[l]{ppl\_norm\_external\_gpt2\_full\\dialogue} & This feature measures normalized perplexity under an external GPT-2 model on the full dialogue stream, capturing overall predictability to a general-domain model. \\
\featureline
\makecell[l]{ppl\_norm\_external\_gpt2\_mean\\scopes} & This feature averages the external GPT-2 perplexity summaries across scoring scopes, providing a broader predictability estimate. \\
\featureline
\midrule
\multicolumn{2}{@{}l}{\textbf{Transcription / Data Quality}} \\
\midrule
partial\_word\_pct & This feature measures the rate of partial-word markers in the transcript, capturing truncations or cut-off productions. \\
\featureline
unintelligible\_marker\_pct & This feature measures the rate of unintelligible markers such as 'xxx' or related annotations, capturing transcription uncertainty or inaudible speech. \\
\featureline
\midrule
\bottomrule
\end{longtable}
\end{scriptsize}

\begin{scriptsize}
\setlength{\tabcolsep}{2.5pt}
\setlength{\LTleft}{0pt}
\setlength{\LTright}{0pt}
\renewcommand{\arraystretch}{1.0}
\renewcommand\theadfont{\bfseries}
\renewcommand\theadalign{cc}
\renewcommand\theadgape{}
\newcommand{\featureline}{\specialrule{0.15pt}{0pt}{0pt}}

\begin{longtable}{
>{\raggedright\arraybackslash}p{0.27\textwidth}
*{9}{>{\centering\arraybackslash}p{0.065\textwidth}}
}
\caption{Full linguistic feature results grouped by category, sorted by Top10 hit count within each category. \textbf{Top10} is the total number of model--target analyses in which a feature appeared in a method's top-10 list; \textbf{Meth.} is the number of methods (Spearman, lasso, XGBoost) for which the feature appeared at least once; \textbf{Spear.}, \textbf{Lasso}, and \textbf{XGB} give the corresponding method-specific hit counts; \textbf{Mean rank} is the average within-method top-10 rank across all hits; \textbf{Mean $|\rho|$} is the mean absolute Spearman correlation across the feature's Spearman top-10 appearances; \textbf{Mean $|\beta|$} is the mean absolute lasso coefficient across the feature's lasso top-10 appearances; \textbf{Mean imp.} is the mean XGBoost feature-importance score across the feature's XGBoost top-10 appearances.}
\label{app_tab:ling_full_grouped}\\
\toprule
\textbf{Feature} & \thead{Top10} & \thead{Meth.} & \thead{Spear.} & \thead{Lasso} & \thead{XGB} & \thead{Mean\\rank} & \thead{Mean\\$|\rho|$} & \thead{Mean\\$|\beta|$} & \thead{Mean\\imp.} \\
\midrule
\endfirsthead
\toprule
\textbf{Feature} & \thead{Top10} & \thead{Meth.} & \thead{Spear.} & \thead{Lasso} & \thead{XGB} & \thead{Mean\\rank} & \thead{Mean\\$|\rho|$} & \thead{Mean\\$|\beta|$} & \thead{Mean\\imp.} \\
\midrule
\endhead
\multicolumn{10}{@{}l}{\textbf{Lexical / Distributional / Scale}}\\
\featureline
bigram\_mutual\_information & 9 & 3 & 3 & 1 & 5 & 5.56 & 0.72 & 0.153 & 0.060 \\
\featureline
hapax\_ratio & 5 & 3 & 1 & 1 & 3 & 6.60 & 0.48 & 0.264 & 0.067 \\
\featureline
trigram\_entropy & 5 & 2 & 4 & 0 & 1 & 5.00 & 0.74 & 0.000 & 0.075 \\
\featureline
num\_conversations & 5 & 1 & 0 & 5 & 0 & 2.00 & 0.00 & 0.178 & 0.000 \\
\featureline
avg\_conversation\_len\_tokens & 4 & 3 & 2 & 1 & 1 & 8.25 & 0.57 & 0.005 & 0.036 \\
\featureline
bigram\_entropy & 4 & 2 & 1 & 0 & 3 & 4.25 & 0.87 & 0.000 & 0.065 \\
\featureline
emotion\_word\_pct & 4 & 2 & 1 & 3 & 0 & 5.75 & 0.40 & 0.013 & 0.000 \\
\featureline
mean\_utterance\_length\_slope & 4 & 2 & 1 & 0 & 3 & 6.00 & 0.50 & 0.000 & 0.052 \\
\featureline
\makecell[l]{conversations\_lt\\512\_tokens\_pct} & 3 & 3 & 1 & 1 & 1 & 7.67 & 0.56 & 0.331 & 0.020 \\
\featureline
num\_types & 3 & 2 & 2 & 1 & 0 & 2.33 & 0.82 & 0.014 & 0.000 \\
\featureline
\makecell[l]{narrative\_discourse\\marker\_pct} & 3 & 2 & 1 & 0 & 2 & 4.33 & 0.48 & 0.000 & 0.044 \\
\featureline
\makecell[l]{syntactic\_complexity\\slope\_over\_time\_proxy} & 3 & 2 & 1 & 0 & 2 & 6.67 & 0.50 & 0.000 & 0.054 \\
\featureline
burstiness\_measure & 3 & 2 & 0 & 1 & 2 & 6.67 & 0.00 & 0.218 & 0.039 \\
\featureline
\makecell[l]{narrative\_discourse\\marker\_token\_pct} & 3 & 2 & 2 & 0 & 1 & 7.33 & 0.43 & 0.000 & 0.014 \\
\featureline
\makecell[l]{frequency\\distribution\_skewness} & 3 & 1 & 3 & 0 & 0 & 4.67 & 0.72 & 0.000 & 0.000 \\
\featureline
intrinsic\_perplexity\_proxy & 3 & 1 & 0 & 0 & 3 & 7.00 & 0.00 & 0.000 & 0.037 \\
\featureline
dis\_legomena\_ratio & 2 & 2 & 0 & 1 & 1 & 6.00 & 0.00 & 0.004 & 0.034 \\
\featureline
elongated\_word\_frequency & 2 & 2 & 0 & 1 & 1 & 7.50 & 0.00 & 0.002 & 0.031 \\
\featureline
eventive\_verb\_token\_pct & 2 & 2 & 0 & 1 & 1 & 8.00 & 0.00 & 0.001 & 0.029 \\
\featureline
num\_tokens & 2 & 2 & 1 & 0 & 1 & 8.50 & 0.84 & 0.000 & 0.022 \\
\featureline
lexical\_diversity\_slope & 2 & 1 & 0 & 0 & 2 & 5.50 & 0.00 & 0.000 & 0.041 \\
\featureline
sound\_effect\_frequency & 1 & 1 & 0 & 0 & 1 & 2.00 & 0.00 & 0.000 & 0.108 \\
\featureline
\makecell[l]{conversations\_lt\\128\_tokens\_pct} & 1 & 1 & 0 & 1 & 0 & 4.00 & 0.00 & 0.002 & 0.000 \\
\featureline
past\_tense\_verb\_pct & 1 & 1 & 0 & 1 & 0 & 4.00 & 0.00 & 0.350 & 0.000 \\
\featureline
num\_utterances & 1 & 1 & 1 & 0 & 0 & 7.00 & 0.85 & 0.000 & 0.000 \\
\featureline
mattr & 1 & 1 & 0 & 0 & 1 & 7.00 & 0.00 & 0.000 & 0.036 \\
\featureline
\makecell[l]{vocabulary\_growth\\over\_time\_slope} & 1 & 1 & 0 & 0 & 1 & 7.00 & 0.00 & 0.000 & 0.039 \\
\featureline
compression\_ratio & 1 & 1 & 1 & 0 & 0 & 9.00 & 0.38 & 0.000 & 0.000 \\
\featureline
past\_tense\_verb\_token\_pct & 1 & 1 & 0 & 0 & 1 & 9.00 & 0.00 & 0.000 & 0.035 \\
\featureline
interjection\_frequency & 1 & 1 & 0 & 0 & 1 & 10.00 & 0.00 & 0.000 & 0.032 \\
\featureline
avg\_sentence\_len\_tokens & 0 & 0 & 0 & 0 & 0 & -- & 0.00 & 0.000 & 0.000 \\
\featureline
avg\_utterance\_len\_tokens & 0 & 0 & 0 & 0 & 0 & -- & 0.00 & 0.000 & 0.000 \\
\featureline
character\_entropy & 0 & 0 & 0 & 0 & 0 & -- & 0.00 & 0.000 & 0.000 \\
\featureline
diminutive\_frequency & 0 & 0 & 0 & 0 & 0 & -- & 0.00 & 0.000 & 0.000 \\
\featureline
eventive\_verb\_pct & 0 & 0 & 0 & 0 & 0 & -- & 0.00 & 0.000 & 0.000 \\
\featureline
num\_sentences & 0 & 0 & 0 & 0 & 0 & -- & 0.00 & 0.000 & 0.000 \\
\featureline
positive\_sentiment\_word\_pct & 0 & 0 & 0 & 0 & 0 & -- & 0.00 & 0.000 & 0.000 \\
\featureline
praise\_word\_pct & 0 & 0 & 0 & 0 & 0 & -- & 0.00 & 0.000 & 0.000 \\
\featureline
token\_entropy & 0 & 0 & 0 & 0 & 0 & -- & 0.00 & 0.000 & 0.000 \\
\featureline
top1000\_token\_coverage & 0 & 0 & 0 & 0 & 0 & -- & 0.00 & 0.000 & 0.000 \\
\featureline
top100\_token\_proportion & 0 & 0 & 0 & 0 & 0 & -- & 0.00 & 0.000 & 0.000 \\
\featureline
ttr & 0 & 0 & 0 & 0 & 0 & -- & 0.00 & 0.000 & 0.000 \\
\featureline
unigram\_entropy & 0 & 0 & 0 & 0 & 0 & -- & 0.00 & 0.000 & 0.000 \\
\featureline
zipf\_slope & 0 & 0 & 0 & 0 & 0 & -- & 0.00 & 0.000 & 0.000 \\
\featureline
\midrule
\multicolumn{10}{@{}l}{\textbf{POS / Syntactic Composition}}\\
\featureline
pos\_num\_tagged\_tokens\_child & 7 & 3 & 1 & 4 & 2 & 2.00 & 0.86 & 0.604 & 0.356 \\
\featureline
\makecell[l]{pos\_num\_tagged\\tokens\_caregiver} & 7 & 3 & 4 & 1 & 2 & 4.29 & 0.74 & 0.016 & 0.062 \\
\featureline
pos\_pos\_bigram\_entropy & 6 & 3 & 1 & 1 & 4 & 3.33 & 0.56 & 0.263 & 0.069 \\
\featureline
\makecell[l]{dep\_total\\utterances\_available} & 5 & 3 & 1 & 1 & 3 & 3.40 & 0.85 & 0.012 & 0.149 \\
\featureline
pos\_pos\_entropy\_caregiver & 4 & 3 & 1 & 1 & 2 & 4.25 & 0.64 & 0.003 & 0.061 \\
\featureline
dep\_num\_sentences & 4 & 2 & 3 & 1 & 0 & 4.75 & 0.70 & 0.530 & 0.000 \\
\featureline
\makecell[l]{pos\_pos\_bigram\\diversity\_tagspace} & 3 & 3 & 1 & 1 & 1 & 4.67 & 0.28 & 0.003 & 0.121 \\
\featureline
pos\_pct\_nouns\_caregiver & 3 & 2 & 1 & 0 & 2 & 2.67 & 0.37 & 0.000 & 0.059 \\
\featureline
dep\_pct\_coordinate\_clauses & 3 & 2 & 1 & 0 & 2 & 4.00 & 0.40 & 0.000 & 0.066 \\
\featureline
pos\_pct\_adverbs & 3 & 2 & 1 & 0 & 2 & 5.00 & 0.45 & 0.000 & 0.029 \\
\featureline
pos\_pos\_entropy & 3 & 2 & 2 & 0 & 1 & 5.33 & 0.49 & 0.000 & 0.073 \\
\featureline
dep\_num\_docs\_parsed & 3 & 2 & 1 & 0 & 2 & 7.33 & 0.49 & 0.000 & 0.027 \\
\featureline
pos\_pos\_bigram\_diversity & 3 & 1 & 3 & 0 & 0 & 9.33 & 0.69 & 0.000 & 0.000 \\
\featureline
dep\_avg\_parse\_tree\_depth & 2 & 2 & 1 & 0 & 1 & 1.50 & 0.57 & 0.000 & 0.085 \\
\featureline
pos\_pos\_entropy\_child & 2 & 2 & 1 & 1 & 0 & 2.50 & 0.60 & 0.440 & 0.000 \\
\featureline
\makecell[l]{dep\_avg\_parse\\tree\_depth\_proxy} & 2 & 2 & 1 & 0 & 1 & 3.00 & 0.57 & 0.000 & 0.058 \\
\featureline
dep\_right\_branching\_ratio & 2 & 2 & 1 & 0 & 1 & 4.00 & 0.44 & 0.000 & 0.025 \\
\featureline
pos\_pct\_conjunctions & 2 & 2 & 1 & 0 & 1 & 4.00 & 0.42 & 0.000 & 0.055 \\
\featureline
dep\_pct\_wh\_questions\_parse & 2 & 2 & 1 & 1 & 0 & 7.50 & 0.32 & 0.001 & 0.000 \\
\featureline
pos\_num\_tagged\_tokens & 2 & 2 & 1 & 0 & 1 & 8.50 & 0.84 & 0.000 & 0.023 \\
\featureline
pos\_pct\_nouns\_child & 2 & 2 & 1 & 0 & 1 & 8.50 & 0.31 & 0.000 & 0.041 \\
\featureline
pos\_pct\_content\_words & 2 & 1 & 0 & 0 & 2 & 7.00 & 0.00 & 0.000 & 0.047 \\
\featureline
pos\_pct\_prepositions & 2 & 1 & 0 & 2 & 0 & 7.00 & 0.00 & 0.268 & 0.000 \\
\featureline
\makecell[l]{dep\_utterances\\used\_for\_parsing} & 1 & 1 & 1 & 0 & 0 & 2.00 & 0.86 & 0.000 & 0.000 \\
\featureline
\makecell[l]{dep\_pct\_passive\\constructions} & 1 & 1 & 1 & 0 & 0 & 3.00 & 0.36 & 0.000 & 0.000 \\
\featureline
\makecell[l]{dep\_avg\_clause\\count\_per\_sentence} & 1 & 1 & 1 & 0 & 0 & 4.00 & 0.53 & 0.000 & 0.000 \\
\featureline
\makecell[l]{dep\_avg\_clause\_dependency\\count\_per\_sentence\_proxy} & 1 & 1 & 1 & 0 & 0 & 5.00 & 0.53 & 0.000 & 0.000 \\
\featureline
dep\_left\_branching\_ratio & 1 & 1 & 0 & 1 & 0 & 5.00 & 0.00 & 0.002 & 0.000 \\
\featureline
pos\_pct\_verbs\_caregiver & 1 & 1 & 0 & 1 & 0 & 7.00 & 0.00 & 0.001 & 0.000 \\
\featureline
\makecell[l]{pos\_total\\utterances\_available} & 1 & 1 & 1 & 0 & 0 & 8.00 & 0.85 & 0.000 & 0.000 \\
\featureline
pos\_pct\_determiners & 1 & 1 & 0 & 0 & 1 & 8.00 & 0.00 & 0.000 & 0.032 \\
\featureline
pos\_pct\_verbs & 1 & 1 & 0 & 1 & 0 & 8.00 & 0.00 & 0.348 & 0.000 \\
\featureline
pos\_pct\_verbs\_child & 1 & 1 & 0 & 0 & 1 & 8.00 & 0.00 & 0.000 & 0.011 \\
\featureline
dep\_pct\_questions\_parse & 1 & 1 & 1 & 0 & 0 & 9.00 & 0.31 & 0.000 & 0.000 \\
\featureline
pos\_utterances\_used\_for\_pos & 1 & 1 & 1 & 0 & 0 & 9.00 & 0.85 & 0.000 & 0.000 \\
\featureline
dep\_avg\_dependency\_length & 1 & 1 & 0 & 0 & 1 & 9.00 & 0.00 & 0.000 & 0.034 \\
\featureline
pos\_pct\_pronouns & 1 & 1 & 1 & 0 & 0 & 10.00 & 0.27 & 0.000 & 0.000 \\
\featureline
dep\_pct\_fragments\_parse & 1 & 1 & 0 & 0 & 1 & 10.00 & 0.00 & 0.000 & 0.016 \\
\featureline
\makecell[l]{dep\_dependency\\length\_variance} & 0 & 0 & 0 & 0 & 0 & -- & 0.00 & 0.000 & 0.000 \\
\featureline
dep\_pct\_imperatives\_parse & 0 & 0 & 0 & 0 & 0 & -- & 0.00 & 0.000 & 0.000 \\
\featureline
dep\_pct\_subordinate\_clauses & 0 & 0 & 0 & 0 & 0 & -- & 0.00 & 0.000 & 0.000 \\
\featureline
\makecell[l]{dep\_pct\_yes\_no\\questions\_parse} & 0 & 0 & 0 & 0 & 0 & -- & 0.00 & 0.000 & 0.000 \\
\featureline
pos\_pct\_adjectives & 0 & 0 & 0 & 0 & 0 & -- & 0.00 & 0.000 & 0.000 \\
\featureline
pos\_pct\_function\_words & 0 & 0 & 0 & 0 & 0 & -- & 0.00 & 0.000 & 0.000 \\
\featureline
pos\_pct\_nouns & 0 & 0 & 0 & 0 & 0 & -- & 0.00 & 0.000 & 0.000 \\
\featureline
\midrule
\multicolumn{10}{@{}l}{\textbf{Semantic Coherence}}\\
\featureline
\makecell[l]{sem\_num\_pairs\\child\_to\_caregiver} & 6 & 3 & 2 & 2 & 2 & 6.17 & 0.80 & 0.005 & 0.050 \\
\featureline
\makecell[l]{sem\_num\_pairs\\caregiver\_to\_child} & 4 & 3 & 2 & 1 & 1 & 3.75 & 0.85 & 0.241 & 0.041 \\
\featureline
\makecell[l]{sem\_total\_adjacent\\pairs\_available} & 4 & 3 & 2 & 1 & 1 & 5.25 & 0.80 & 0.762 & 0.090 \\
\featureline
\makecell[l]{sem\_mean\_adjacent\_turn\\semantic\_similarity} & 3 & 3 & 1 & 1 & 1 & 5.67 & 0.43 & 0.001 & 0.044 \\
\featureline
sem\_num\_adjacent\_pairs\_used & 3 & 2 & 2 & 1 & 0 & 6.00 & 0.79 & 0.008 & 0.000 \\
\featureline
\makecell[l]{sem\_mean\_adjacent\_turn\_semantic\\similarity\_len\_weighted} & 1 & 1 & 1 & 0 & 0 & 5.00 & 0.41 & 0.000 & 0.000 \\
\featureline
\makecell[l]{sem\_mean\_semantic\_similarity\\caregiver\_to\_child} & 1 & 1 & 1 & 0 & 0 & 7.00 & 0.51 & 0.000 & 0.000 \\
\featureline
\makecell[l]{sem\_mean\_semantic\_similarity\\child\_to\_caregiver} & 1 & 1 & 1 & 0 & 0 & 8.00 & 0.39 & 0.000 & 0.000 \\
\featureline
\makecell[l]{sem\_total\\utterances\_available} & 1 & 1 & 1 & 0 & 0 & 10.00 & 0.85 & 0.000 & 0.000 \\
\featureline
\makecell[l]{sem\_utterances\\used\_for\_semantic} & 0 & 0 & 0 & 0 & 0 & -- & 0.00 & 0.000 & 0.000 \\
\featureline
\midrule
\multicolumn{10}{@{}l}{\textbf{Mixture / Composition Metadata}}\\
\featureline
div\_mean\_kl\_from\_others & 5 & 3 & 3 & 1 & 1 & 3.60 & 0.71 & 0.500 & 0.053 \\
\featureline
\makecell[l]{mix\_inter\_family\\variance\_entropy} & 4 & 3 & 1 & 1 & 2 & 5.50 & 0.60 & 0.074 & 0.056 \\
\featureline
\makecell[l]{mix\_cross\_family\\lexical\_overlap} & 4 & 2 & 3 & 0 & 1 & 5.50 & 0.49 & 0.000 & 0.078 \\
\featureline
mix\_family\_diversity\_index & 4 & 2 & 2 & 2 & 0 & 5.50 & 0.42 & 0.013 & 0.000 \\
\featureline
div\_median\_js\_with\_others & 3 & 3 & 1 & 1 & 1 & 3.33 & 0.86 & 0.536 & 0.041 \\
\featureline
age\_weighted\_mean\_age\_months & 3 & 2 & 2 & 1 & 0 & 6.33 & 0.31 & 0.013 & 0.000 \\
\featureline
div\_mean\_js\_with\_others & 3 & 1 & 0 & 0 & 3 & 4.00 & 0.00 & 0.000 & 0.080 \\
\featureline
\makecell[l]{mix\_content\_vocab\\support\_size} & 2 & 2 & 1 & 0 & 1 & 1.50 & 0.85 & 0.000 & 0.332 \\
\featureline
\makecell[l]{mix\_mixture\\top100\_token\_mass} & 2 & 2 & 1 & 0 & 1 & 1.50 & 0.64 & 0.000 & 0.170 \\
\featureline
\makecell[l]{mix\_effective\\vocab\_size\_bits} & 2 & 2 & 1 & 0 & 1 & 2.00 & 0.61 & 0.000 & 0.236 \\
\featureline
div\_mean\_kl\_to\_others & 2 & 2 & 1 & 0 & 1 & 3.50 & 0.42 & 0.000 & 0.037 \\
\featureline
age\_num\_component\_families & 2 & 2 & 1 & 0 & 1 & 5.00 & 0.54 & 0.000 & 0.058 \\
\featureline
mix\_mixture\_top10\_token\_mass & 1 & 1 & 0 & 0 & 1 & 1.00 & 0.00 & 0.000 & 0.106 \\
\featureline
mix\_mixture\_top50\_token\_mass & 1 & 1 & 1 & 0 & 0 & 2.00 & 0.62 & 0.000 & 0.000 \\
\featureline
\makecell[l]{mix\_mixture\_vocab\\support\_size} & 1 & 1 & 1 & 0 & 0 & 3.00 & 0.85 & 0.000 & 0.000 \\
\featureline
\makecell[l]{mix\_effective\\vocab\_size\_nats} & 1 & 1 & 1 & 0 & 0 & 4.00 & 0.61 & 0.000 & 0.000 \\
\featureline
mix\_vocab\_support\_size & 1 & 1 & 1 & 0 & 0 & 4.00 & 0.85 & 0.000 & 0.000 \\
\featureline
\makecell[l]{mix\_effective\\vocabulary\_size} & 1 & 1 & 1 & 0 & 0 & 5.00 & 0.61 & 0.000 & 0.000 \\
\featureline
\makecell[l]{mix\_inter\_family\_variance\\syntactic\_complexity} & 1 & 1 & 0 & 0 & 1 & 5.00 & 0.00 & 0.000 & 0.045 \\
\featureline
\makecell[l]{mix\_mixture\_token\\entropy\_bits} & 1 & 1 & 1 & 0 & 0 & 6.00 & 0.61 & 0.000 & 0.000 \\
\featureline
mix\_formatting\_token\_mass & 1 & 1 & 0 & 0 & 1 & 6.00 & 0.00 & 0.000 & 0.051 \\
\featureline
\makecell[l]{mix\_mixture\_token\\entropy\_nats} & 1 & 1 & 1 & 0 & 0 & 7.00 & 0.61 & 0.000 & 0.000 \\
\featureline
mix\_mixture\_top5\_token\_mass & 1 & 1 & 0 & 0 & 1 & 7.00 & 0.00 & 0.000 & 0.034 \\
\featureline
\makecell[l]{age\_age\_metadata\\alignment\_score} & 1 & 1 & 0 & 0 & 1 & 8.00 & 0.00 & 0.000 & 0.039 \\
\featureline
div\_max\_js\_with\_others & 1 & 1 & 0 & 0 & 1 & 8.00 & 0.00 & 0.000 & 0.035 \\
\featureline
age\_num\_families\_with\_age & 1 & 1 & 1 & 0 & 0 & 9.00 & 0.54 & 0.000 & 0.000 \\
\featureline
mix\_num\_component\_families & 1 & 1 & 1 & 0 & 0 & 10.00 & 0.54 & 0.000 & 0.000 \\
\featureline
age\_weighted\_age\_std\_months & 0 & 0 & 0 & 0 & 0 & -- & 0.00 & 0.000 & 0.000 \\
\featureline
\makecell[l]{mix\_effective\_vocab\\size\_content\_only} & 0 & 0 & 0 & 0 & 0 & -- & 0.00 & 0.000 & 0.000 \\
\featureline
\makecell[l]{mix\_js\_divergence\\within\_mixture} & 0 & 0 & 0 & 0 & 0 & -- & 0.00 & 0.000 & 0.000 \\
\featureline
mix\_mixture\_top1\_token\_mass & 0 & 0 & 0 & 0 & 0 & -- & 0.00 & 0.000 & 0.000 \\
\featureline
\makecell[l]{mix\_mixture\_weight\\entropy\_bits} & 0 & 0 & 0 & 0 & 0 & -- & 0.00 & 0.000 & 0.000 \\
\featureline
\makecell[l]{mix\_token\_entropy\\bits\_content\_only} & 0 & 0 & 0 & 0 & 0 & -- & 0.00 & 0.000 & 0.000 \\
\featureline
\midrule
\multicolumn{10}{@{}l}{\textbf{Conversational / Discourse Structure}}\\
\featureline
\makecell[l]{topic\_persistence\_length\\overlap\_heuristic\_proxy} & 4 & 3 & 1 & 2 & 1 & 3.75 & 0.51 & 0.002 & 0.059 \\
\featureline
avg\_conversation\_len\_turns & 3 & 3 & 1 & 1 & 1 & 4.00 & 0.59 & 0.432 & 0.077 \\
\featureline
speaker\_label\_consistency & 3 & 3 & 1 & 1 & 1 & 5.00 & 0.59 & 0.003 & 0.019 \\
\featureline
self\_repair\_frequency & 3 & 3 & 1 & 1 & 1 & 5.33 & 0.54 & 0.353 & 0.066 \\
\featureline
\makecell[l]{overlapping\\speech\_marker\_pct} & 3 & 2 & 1 & 2 & 0 & 5.00 & 0.50 & 0.015 & 0.000 \\
\featureline
child\_token\_pct & 3 & 2 & 0 & 2 & 1 & 6.67 & 0.00 & 0.184 & 0.019 \\
\featureline
fragment\_pct & 2 & 2 & 1 & 0 & 1 & 6.00 & 0.54 & 0.000 & 0.033 \\
\featureline
anaphora\_usage\_pct & 2 & 2 & 1 & 0 & 1 & 7.00 & 0.53 & 0.000 & 0.035 \\
\featureline
turn\_taking\_frequency & 2 & 1 & 0 & 0 & 2 & 7.00 & 0.00 & 0.000 & 0.039 \\
\featureline
caregiver\_child\_token\_ratio & 1 & 1 & 0 & 1 & 0 & 1.00 & 0.00 & 0.838 & 0.000 \\
\featureline
\makecell[l]{unknown\_speaker\\utterance\_pct} & 1 & 1 & 1 & 0 & 0 & 2.00 & 0.59 & 0.000 & 0.000 \\
\featureline
directive\_pct & 1 & 1 & 0 & 1 & 0 & 2.00 & 0.00 & 0.004 & 0.000 \\
\featureline
caregiver\_token\_pct & 1 & 1 & 0 & 1 & 0 & 3.00 & 0.00 & 0.004 & 0.000 \\
\featureline
wh\_question\_pct & 1 & 1 & 1 & 0 & 0 & 4.00 & 0.36 & 0.000 & 0.000 \\
\featureline
caregiver\_question\_pct & 1 & 1 & 1 & 0 & 0 & 6.00 & 0.34 & 0.000 & 0.000 \\
\featureline
expansion\_pct & 1 & 1 & 0 & 1 & 0 & 6.00 & 0.00 & 0.090 & 0.000 \\
\featureline
question\_pct & 1 & 1 & 1 & 0 & 0 & 7.00 & 0.34 & 0.000 & 0.000 \\
\featureline
\makecell[l]{caregiver\_after\\child\_any\_overlap\_pct} & 1 & 1 & 1 & 0 & 0 & 8.00 & 0.49 & 0.000 & 0.000 \\
\featureline
\makecell[l]{caregiver\_reuse\\child\_tokens\_pct} & 1 & 1 & 1 & 0 & 0 & 9.00 & 0.49 & 0.000 & 0.000 \\
\featureline
pronoun\_density & 1 & 1 & 1 & 0 & 0 & 9.00 & 0.27 & 0.000 & 0.000 \\
\featureline
imperative\_pct & 1 & 1 & 0 & 1 & 0 & 9.00 & 0.00 & 0.000 & 0.000 \\
\featureline
avg\_child\_utterance\_len & 1 & 1 & 1 & 0 & 0 & 10.00 & 0.49 & 0.000 & 0.000 \\
\featureline
\makecell[l]{adjacent\_turn\\lexical\_overlap} & 1 & 1 & 0 & 1 & 0 & 10.00 & 0.00 & 0.000 & 0.000 \\
\featureline
avg\_caregiver\_utterance\_len & 0 & 0 & 0 & 0 & 0 & -- & 0.00 & 0.000 & 0.000 \\
\featureline
caregiver\_wh\_question\_pct & 0 & 0 & 0 & 0 & 0 & -- & 0.00 & 0.000 & 0.000 \\
\featureline
clarification\_pct & 0 & 0 & 0 & 0 & 0 & -- & 0.00 & 0.000 & 0.000 \\
\featureline
confirmation\_pct & 0 & 0 & 0 & 0 & 0 & -- & 0.00 & 0.000 & 0.000 \\
\featureline
\makecell[l]{topic\_entropy\_overlap\\heuristic\_proxy} & 0 & 0 & 0 & 0 & 0 & -- & 0.00 & 0.000 & 0.000 \\
\featureline
yes\_no\_question\_pct & 0 & 0 & 0 & 0 & 0 & -- & 0.00 & 0.000 & 0.000 \\
\featureline
\midrule
\multicolumn{10}{@{}l}{\textbf{Repetition / Formulaicity}}\\
\featureline
repeated\_trigram\_ratio & 4 & 3 & 1 & 1 & 2 & 5.25 & 0.37 & 0.004 & 0.085 \\
\featureline
ngram\_repetition\_rate & 4 & 2 & 1 & 0 & 3 & 2.75 & 0.31 & 0.000 & 0.116 \\
\featureline
\makecell[l]{exact\_sentence\\repetition\_rate} & 2 & 2 & 1 & 0 & 1 & 4.00 & 0.50 & 0.000 & 0.145 \\
\featureline
repeated\_bigram\_ratio & 2 & 1 & 0 & 0 & 2 & 5.00 & 0.00 & 0.000 & 0.118 \\
\featureline
partial\_repetition\_frequency & 1 & 1 & 0 & 1 & 0 & 8.00 & 0.00 & 0.011 & 0.000 \\
\featureline
reduplication\_frequency & 0 & 0 & 0 & 0 & 0 & -- & 0.00 & 0.000 & 0.000 \\
\featureline
\midrule
\multicolumn{10}{@{}l}{\textbf{Reference LM Predictability}}\\
\featureline
\makecell[l]{ppl\_norm\_babylm20m\\full\_dialogue} & 3 & 2 & 0 & 1 & 2 & 6.33 & 0.00 & 0.012 & 0.050 \\
\featureline
\makecell[l]{ppl\_norm\_babylm20m\\caregiver\_only} & 2 & 1 & 0 & 2 & 0 & 6.50 & 0.00 & 0.003 & 0.000 \\
\featureline
\makecell[l]{ppl\_norm\_babylm20m\\mean\_scopes} & 2 & 1 & 0 & 0 & 2 & 9.50 & 0.00 & 0.000 & 0.026 \\
\featureline
\makecell[l]{ppl\_norm\_external\\gpt2\_mean\_scopes} & 1 & 1 & 0 & 0 & 1 & 10.00 & 0.00 & 0.000 & 0.019 \\
\featureline
\makecell[l]{ppl\_norm\_external\\gpt2\_caregiver\_only} & 0 & 0 & 0 & 0 & 0 & -- & 0.00 & 0.000 & 0.000 \\
\featureline
\makecell[l]{ppl\_norm\_external\\gpt2\_full\_dialogue} & 0 & 0 & 0 & 0 & 0 & -- & 0.00 & 0.000 & 0.000 \\
\featureline
\midrule
\multicolumn{10}{@{}l}{\textbf{Transcription / Data Quality}}\\
\featureline
unintelligible\_marker\_pct & 2 & 2 & 0 & 1 & 1 & 9.00 & 0.00 & 0.001 & 0.031 \\
\featureline
partial\_word\_pct & 0 & 0 & 0 & 0 & 0 & -- & 0.00 & 0.000 & 0.000 \\
\featureline
\bottomrule
\end{longtable}
\end{scriptsize}

\end{document}